\documentclass{article}

\PassOptionsToPackage{numbers,sort&compress}{natbib}

\usepackage[preprint]{neurips_2026}

\usepackage[utf8]{inputenc}
\usepackage[T1]{fontenc}
\usepackage{amsmath,amssymb,amsfonts}
\usepackage{booktabs}
\usepackage{graphicx}
\usepackage{hyperref}
\usepackage[ocgcolorlinks]{ocgx2} %this forces colours in the PDF
\usepackage{cleveref}
\usepackage{microtype}
\usepackage{xcolor}
\usepackage{enumitem}
\usepackage{pifont}
\usepackage{tikz}
\usetikzlibrary{arrows.meta, positioning, shapes.geometric, fit, calc, backgrounds, decorations.pathreplacing}
\usepackage{diagbox}

\graphicspath{{figures/}}

\title{Cycle-Consistent Neural Explanation \\ of Formal Verification Certificates}

\author{
  Andoni Rodriguez, Alberto Pozanco, Daniel Borrajo\\
  J.P. Morgan AI Research
}

\date{}

\begin{document}
\maketitle

\begin{abstract}
Formal verification produces machine-checkable certificates that attest to the satisfaction or violation of temporal properties, yet these certificates remain opaque to non-specialist stakeholders. We propose a cycle-consistent neural architecture that generates faithful natural language explanations of verification certificates. A forward network $\mathrm{NN}_1$ maps certificates to explanations, and an inverse network $\mathrm{NN}_2$ reconstructs certificates from explanations; a symbolic verifier closes the loop, providing a differentiable faithfulness proxy. A pointer-generator mechanism ensures lexical grounding by copying state names directly from the certificate.

We evaluate on 420 test certificates spanning six verification methods (bounded proof, $k$-induction, inductive invariant, lasso, reachability, witness pair) in both YES and NO verdict variants, drawn from a financial compliance domain with 207 named states. Our trained architecture, combined with a hybrid inference-time routing strategy, achieves 90.0\% cycle-verified soundness---surpassing a multi-LLM few-shot baseline (76.1\% for the best of 16 LLM combinations across four frontier models) by 13.9 percentage points. The neural model wins on 10 of 12 verdict/kind categories, with three categories reaching 100\% soundness. The architecture offers ${\sim}860\times$ faster inference (${\sim}185$\,ms vs.\ ${\sim}160$\,s per certificate for the full multi-LLM baseline), offline operation, deterministic outputs, and zero per-inference cost. These results demonstrate that trained specialization outperforms general-purpose LLM prompting for structured certificate explanation, while eliminating the deployment constraints of cloud-based inference.
\end{abstract}

%% ---- Sections ----
\section{Introduction}
\label{sec:introduction}

Formal verification produces
machine-checkable certificates~\cite{biere1999symbolic,
clarke2018model,baier2008principles}---lasso-shaped counterexamples, reachability
witnesses, inductive invariants, bounded proofs---that provide mathematical guarantees about system
behavior. In regulated domains such as financial compliance, these certificates are the evidentiary basis for regulatory assurance, yet they remain opaque to the auditors and compliance officers who must act on them. Generating natural language explanations that are simultaneously \emph{faithful} (no hallucinated facts) and \emph{fluent} (comprehensible to non-specialists) is an open challenge: existing neural approaches prioritize fluency at the expense of grounding, while template systems are rigid and domain-specific.

We observe that faithfulness admits a testable formulation: an explanation~$E$ of a certificate~$C$ is faithful if and only if~$C$ can be reconstructed from~$E$ alone. This motivates a \emph{cycle-consistent} architecture: a forward pointer-generator network $\mathrm{NN}_1$ maps $C \rightarrow E$, copying state names directly from the certificate to eliminate entity hallucination; an inverse Transformer $\mathrm{NN}_2$ reconstructs $E \rightarrow C'$; and a symbolic verifier compares $C$ with $C'$, providing a faithfulness proxy that is made differentiable through surrogate reconstruction losses (Section~\ref{sec:cycle}), requiring no human annotation. In addition, a hybrid inference-time router selects among several decoding configurations per example, exploiting structural differences across certificate types.

We evaluate on 420 test certificates spanning 12 verdict/kind categories from financial compliance workflows. The hybrid router achieves 90.0\% cycle-verified soundness---a lower bound on true faithfulness, since NN$_2$'s 3.6\% false-negative rate means some genuinely faithful explanations are conservatively flagged. This surpasses the best of 16 frontier LLM pair configurations (76.1\%) by 13.9 percentage points (non-overlapping 95\% CIs), winning 10 of 12 categories with three at 100\% and no losses. The sub-1M parameter model operates offline at ${\sim}860\times$ lower latency, with deterministic outputs, zero per-inference cost, and no data exposure to external services---properties critical for regulated industries. Our contributions:

\begin{enumerate}
    \item \textbf{Cycle-consistent explanation.} A neural architecture combining a pointer-generator forward model, an inverse Transformer, and a symbolic verifier to enforce faithfulness via cycle consistency without human annotation (Section~\ref{sec:methodology}).

    \item \textbf{Hybrid inference-time routing.} A test-time strategy achieving 90.0\% soundness vs.\ 61.7\% for the best single configuration, requiring no additional training (Section~\ref{sec:routing_results}).

    \item \textbf{Surpassing LLM prompting.} A sub-1M parameter model outperforms 16 frontier LLM configurations on cycle-verified soundness (90.0\% vs.\ 76.1\%), providing evidence that trained specialization outperforms general-purpose prompting when the faithfulness criterion is formally verifiable (Section~\ref{sec:baseline}).

    \item \textbf{Deployment-ready architecture.} Offline operation, deterministic outputs, zero marginal cost, and no data exposure---with a fixed computation graph that lacks the free-text input channel through which prompt injection attacks operate (Section~\ref{sec:experiments}).
\end{enumerate}

\section{Related Work}
\label{sec:related_work}

Our approach sits at the intersection of four research areas, which we discuss in turn before positioning our contribution against prior work (Table~\ref{tab:related_positioning}).

\paragraph{Cycle-consistent learning.}
Cycle consistency as a training objective was popularized by CycleGAN~\cite{zhu2017unpaired} for unpaired image-to-image translation, enforcing that a mapping $G\colon X \rightarrow Y$ composed with its inverse $F\colon Y \rightarrow X$ approximates the identity. In natural language processing, back-translation~\cite{sennrich2016improving} applies an analogous principle: translating a sentence to a pivot language and back, then penalizing divergence from the original. Subsequent work extended cycle consistency to style transfer~\cite{prabhumoye2018style}, data augmentation~\cite{fadaee2017data}, and unsupervised machine translation~\cite{lample2018unsupervised}. Our work differs in a fundamental respect: the cycle traverses two \emph{different modalities}---structured symbolic certificates and natural language text---rather than two instances of the same modality (e.g., two image domains or two languages). Moreover, the faithfulness criterion at the cycle's endpoint is not approximate reconstruction but \emph{symbolic semantic equivalence}, evaluated by a formal verifier that compares state sets, path structures, and component memberships.

\paragraph{LLM reward signals and judging.}
The use of large language models to provide training-time signals has gained traction through RLAIF~\cite{bai2022constitutional}, LLM-as-Judge~\cite{zheng2023judging}, and Constitutional AI~\cite{bai2022constitutional}. In these frameworks, the LLM typically judges \emph{overall quality}---a single signal conflating correctness, fluency, helpfulness, and safety. Our architecture introduces a critical separation: the LLM judges \emph{only fluency}, while faithfulness is enforced by an entirely independent mechanism (cycle consistency with symbolic verification), ensuring that fluency optimization cannot introduce hallucinations.

Our hybrid router is also related to test-time compute scaling, where generating and reranking multiple candidates improves output quality~\cite{brown2024large}; our twist is that the reranking oracle is an \emph{exact symbolic verifier} rather than a learned reward model, and the sub-1M model makes evaluating dozens of candidates per certificate computationally trivial.

\paragraph{Neural text generation from structured data.}
Data-to-text generation maps structured inputs to natural language. Early neural approaches used sequence-to-sequence models with attention~\cite{bahdanau2015neural}, later augmented with copy mechanisms~\cite{gulcehre2016pointing,see2017get}. Recent work includes structured data verbalization~\cite{chen2020logical}, table-to-text with pre-trained models~\cite{parikh2020totto}, and knowledge-grounded generation~\cite{ji2023survey}. We adapt pointer-generator networks~\cite{see2017get} to certificate-to-explanation generation; our setting is distinguished by formally verifiable faithfulness and the finding that a small trained model surpasses LLM few-shot prompting (90.0\% vs.\ 76.1\%).

A parallel line of work measures generation faithfulness post hoc, using
learned proxies: entailment-based factual-consistency classifiers~\cite{kryscinski2020evaluating},
question-answering-based consistency metrics~\cite{durmus2020feqa,wang2020asking},
and the broader hallucination taxonomies surveyed by Ji et al.~\cite{ji2023survey}.
These metrics approximate faithfulness with a trained or prompted scorer and are
therefore subject to their own errors. Our criterion is fundamentally different:
faithfulness is not estimated but \emph{exactly verified}, since the symbolic
verifier checks set- and sequence-level equivalence between the original and
reconstructed certificate. This exactness is possible only because certificates
are structured objects with a well-defined fact set.

\paragraph{Explainability in formal verification.}
Counterexample explanation has been addressed through simulation~\cite{groce2004understanding}, abstraction-refinement traces~\cite{clarke2000counterexample} and fault localization~\cite{beer2012explaining}. In general, template-based systems are faithful by construction but fundamentally inflexible---each new certificate kind or domain requires manual template authoring. Our neural approach addresses these limitations through learned generation, while maintaining faithfulness enforcement via cycle consistency rather than template rigidity.

In the realm of very recent work in reactive systems, \cite{bassan23formally} produce formally-guaranteed explanations of reactive neural networks via abductive formal reasoning and \cite{rodriguez25explanations} generate simple counter-examples (similar to the traces we use) to assess a system is not realizable. Neither of these approaches produce NL explanations whose formalism in double checked with a cycle consistent architecture.

\paragraph{Neuro-symbolic and verifier-in-the-loop generation.}
A growing body of work couples neural generators with symbolic or
programmatic checkers that constrain or validate outputs, spanning
verifier-guided decoding, tool- and checker-augmented generation, and
neuro-symbolic NLG that enforces logical constraints on generated
text~\cite{nye2021improving,poesia2022synchromesh}. A closely related but
\emph{inverse} task is autoformalization, which maps natural language into
formal statements---e.g., translating informal mathematics or specifications
into a proof assistant or temporal logic~\cite{wu2022autoformalization}. Our
forward task is the opposite direction: we \emph{de-formalize}, mapping a
formal certificate to natural language. The connection is that our inverse
network $\mathrm{NN}_2$ performs a constrained, domain-specific form of
autoformalization (explanation $\rightarrow$ certificate), so the cycle as a
whole composes verbalization with re-formalization under a symbolic verifier.
Unlike autoformalization systems, however, faithfulness here is not the
correctness of a single NL$\rightarrow$formal translation but the
\emph{round-trip equivalence} of the original and reconstructed certificate.
More broadly, our architecture is verifier-in-the-loop in a strong sense: the
symbolic verifier is not merely a post-hoc filter but closes a differentiable
training cycle, and at inference time it serves as the selection oracle for
routing. Unlike systems where the checker validates surface form, our verifier
evaluates \emph{semantic equivalence} of reconstructed certificates, making
faithfulness a property the architecture optimizes rather than one it audits
after the fact.

\paragraph{Positioning.}
The novelty lies not in any single component but in their \emph{principled integration} for a domain where faithfulness is formally verifiable. The separation of fluency from faithfulness provides a structural property absent from prior neural explanation systems: the faithfulness constraint is enforced by a symbolic verifier, not by a learned or prompted model. Table~\ref{tab:related_positioning} summarizes how our approach compares to prior work along the dimensions of faithfulness, fluency, supervision requirements, and formal grounding.

\begin{table}[h]
\centering
\caption{Comparison with related approaches. Our work uniquely combines cycle-consistent faithfulness, LLM fluency scoring, and pointer-generator copying for formal verification certificates.}
\label{tab:related_positioning}
\small
\begin{tabular}{lcccc}
\toprule
\textbf{Approach} & \textbf{Faithful} & \textbf{Fluent} & \textbf{Unsupervised} & \textbf{Formal} \\
\midrule
Template systems & \checkmark & $\times$ & $\times$ & \checkmark \\
CycleGAN / Back-translation & $\sim$ & $\sim$ & \checkmark & $\times$ \\
RLAIF / LLM-as-Judge & $\times$ & \checkmark & $\sim$ & $\times$ \\
Data-to-text (supervised) & $\sim$ & \checkmark & $\times$ & $\times$ \\
\textbf{Ours} & \checkmark & \checkmark & $\times$ & \checkmark \\
\bottomrule
\end{tabular}
\end{table}
\section{Methodology}
\label{sec:methodology}

We present a cycle-consistent neural architecture for generating natural language explanations of formal verification certificates. The system comprises two jointly trained neural networks, a symbolic verifier, and a hybrid inference-time routing strategy (Figure~\ref{fig:architecture}).

\subsection{Problem Formulation}
\label{sec:problem}

Let $\mathcal{C}$ denote the space of verification certificates and $\mathcal{E}$ the space of natural language explanations. A certificate $C \in \mathcal{C}$ is a structured object comprising a verdict (YES/NO), a kind (reachability, lasso, witness\_pair, inductive\_invariant, bounded\_proof, $k$-induction), and kind-specific content (state sequences, path structures, invariant sets). An explanation $E \in \mathcal{E}$ is a natural language sentence or paragraph describing the certificate's meaning.
The underlying temporal properties are specified in standard logics such as
LTL~\cite{pnueli1977temporal} and CTL~\cite{clarke1986automatic}; our method
consumes the resulting certificates and is agnostic to the source logic.

We seek a mapping $f\colon \mathcal{C} \rightarrow \mathcal{E}$ that produces explanations satisfying two properties:

\begin{itemize}
    \item \textbf{Soundness (no hallucination).} Every factual claim in $E$ is supported by $C$. If the explanation mentions a state~$s$, then~$s$ must appear in the certificate. If it claims a path exists, that path must be present in~$C$. Soundness is a hard constraint: any violation constitutes a hallucination that could mislead a human decision-maker.

    \item \textbf{Fluency.} The explanation reads as natural English comprehensible to domain experts who are not formal methods specialists.
\end{itemize}

We additionally define a desirable but certificate-type-dependent property:

\begin{itemize}
    \item \textbf{Completeness (coverage).} The explanation mentions a sufficient proportion of the certificate's essential content. Unlike soundness, completeness is not a binary correctness criterion but an \emph{editorial quality}: the appropriate level of completeness varies by certificate type and intended audience. An exhaustive listing of every state in a 10-step reachability path may reduce clarity, while omitting states from an inductive invariant may undermine the proof's validity.
\end{itemize}

This distinction between soundness and completeness is fundamental to our approach. Soundness must be maximized universally. Completeness is a learned editorial judgment that the model acquires from training examples.

\subsection{Faithfulness as Cycle Consistency}
\label{sec:cycle}

\begin{figure}[t]
\centering
\resizebox{\columnwidth}{!}{%
\begin{tikzpicture}[
    node distance=1.4cm and 1.6cm,
    block/.style={rectangle, draw, rounded corners=3pt, minimum height=0.9cm, minimum width=1.5cm, align=center, font=\small},
    data/.style={rectangle, draw, minimum height=0.7cm, minimum width=1.3cm, align=center, font=\small, fill=blue!8},
    loss/.style={rectangle, draw, dashed, minimum height=0.6cm, minimum width=1.2cm, align=center, font=\footnotesize, fill=red!8},
    arr/.style={-{Stealth[length=2.5mm]}, thick},
    darr/.style={-{Stealth[length=2.5mm]}, thick, dashed, red!70!black},
]

% Main data flow
\node[data] (C) {$C$\\\footnotesize Certificate};
\node[block, right=of C, fill=green!12] (NN1) {$\mathrm{NN}_1$\\\footnotesize Pointer-Gen};
\node[data, right=of NN1] (E) {$E$\\\footnotesize Explanation};
\node[block, right=of E, fill=orange!12] (NN2) {$\mathrm{NN}_2$\\\footnotesize Transformer};
\node[data, right=of NN2] (Cprime) {$C'$\\\footnotesize Reconstructed};

% Arrows for forward pass
\draw[arr] (C) -- (NN1);
\draw[arr] (NN1) -- (E);
\draw[arr] (E) -- (NN2);
\draw[arr] (NN2) -- (Cprime);

% Symbolic verifier
\node[block, below=1.2cm of $(E)$, fill=yellow!15, minimum width=2.2cm] (V) {Symbolic Verifier\\\footnotesize $\mathrm{V}(C, C')$};

% Verifier connections
\draw[arr, gray] (C.south) -- ++(0,-0.4) -| (V.north west);
\draw[arr, gray] (Cprime.south) -- ++(0,-0.4) -| (V.north east);

% Loss signals
\node[loss, below=0.5cm of V] (Lcycle) {$\mathcal{L}_{\mathrm{cycle}}$};
\draw[darr] (V) -- (Lcycle);

% Recon loss
\node[loss, above=0.5cm of NN1] (Lrecon) {$\mathcal{L}_{\mathrm{recon}}$};
\node[data, above=0.5cm of E, minimum width=1.3cm, fill=blue!15] (Eref) {$E^{\mathrm{ref}}$\\\footnotesize Reference};
\draw[arr, gray] (Eref.west) -| (Lrecon.east);
\draw[darr] (Lrecon) -- (NN1);

% Copy mechanism annotation
\draw[arr, green!50!black, thick, bend left=40] (C.north) to node[above, font=\scriptsize, text=green!40!black] {copy} (NN1.north);

\end{tikzpicture}%
}
\caption{Cycle-consistent architecture. Certificate~$C$ is mapped to explanation~$E$ by $\mathrm{NN}_1$ (pointer-generator), then reconstructed as~$C'$ by $\mathrm{NN}_2$ (Transformer). A symbolic verifier compares $C$ and $C'$, producing a cycle consistency loss~$\mathcal{L}_{\mathrm{cycle}}$. The pointer-generator copies state names directly from the encoded certificate, while generating English scaffolding from the vocabulary.}
\label{fig:architecture}
\end{figure}
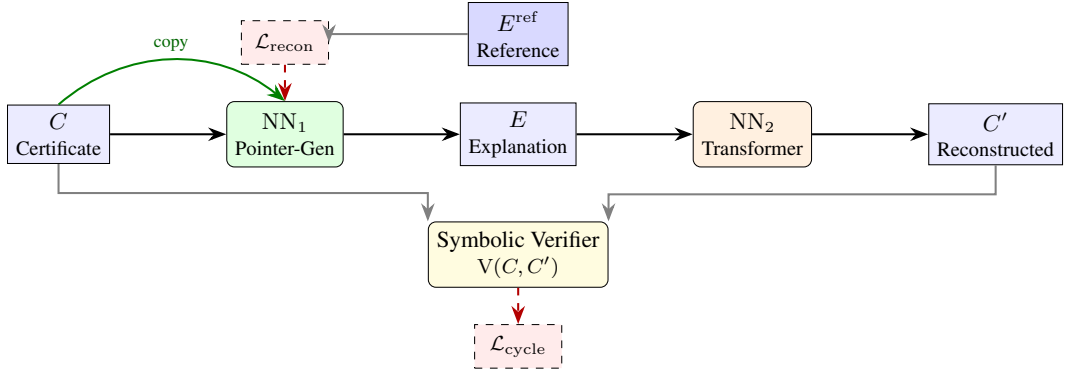

We operationalize soundness through a cycle-consistent architecture (see Figure~\ref{fig:architecture}). A forward model $\mathrm{NN}_1\colon \mathcal{C} \rightarrow \mathcal{E}$ generates explanations, and an inverse model $\mathrm{NN}_2\colon \mathcal{E} \rightarrow \mathcal{C}$ reconstructs certificates from explanations. Given a certificate~$C$:

\begin{equation}
    E = \mathrm{NN}_1(C), \qquad C' = \mathrm{NN}_2(E)
\end{equation}

Soundness is enforced by requiring that the reconstructed certificate~$C'$ is consistent with the original~$C$. We evaluate this through a symbolic verifier $\mathrm{V}(C, C')$ that decomposes the comparison into:

\begin{equation}
    \text{Soundness}(C, C') = \frac{|\mathcal{F}(C) \cap \mathcal{F}(C')|}{|\mathcal{F}(C')|}, \qquad
    \text{Completeness}(C, C') = \frac{|\mathcal{F}(C) \cap \mathcal{F}(C')|}{|\mathcal{F}(C)|}
    \label{eq:soundness}
\end{equation}

where $\mathcal{F}(\cdot)$ extracts the set of facts---states, path segments, and structural elements---from a certificate. Soundness measures precision (of the facts $\mathrm{NN}_2$ reconstructed, how many were in the original?), and completeness measures recall (of the facts in the original, how many survived the cycle?). In the evaluation metrics of Section~\ref{sec:metrics}, we report completeness under the name \emph{coverage}.

The cycle consistency loss used during training approximates this check differentiably:

\begin{equation}
    \mathcal{L}_{\mathrm{cycle}} = \mathcal{L}_{\mathrm{cycle\text{-}recon}}(C, C') + \mathcal{L}_{\mathrm{cycle\text{-}sem}}(C, C')
\end{equation}

where $\mathcal{L}_{\mathrm{cycle\text{-}recon}}$ is the token-level reconstruction loss between the encoded representations of~$C$ and~$C'$, and $\mathcal{L}_{\mathrm{cycle\text{-}sem}}$ is the semantic divergence measured by the symbolic verifier.

Crucially, the cycle enforces soundness by construction: if $\mathrm{NN}_1$ hallucinates a state that is not in~$C$, then $\mathrm{NN}_2$ will reconstruct a certificate containing that hallucinated state, and the verifier will penalize the mismatch. The only way to minimize cycle loss is for $\mathrm{NN}_1$ to produce explanations whose content is faithfully grounded in the input certificate. 

The appropriate level of completeness varies by certificate type---reachability witnesses tolerate lower completeness while inductive invariants require high completeness---and is learned from training examples rather than hard-coded (see Appendix~\ref{app:completeness} for details).

\subsection{Architecture}
\label{sec:architecture}

\paragraph{Certificate Encoder.} Certificates are encoded as token sequences with a structured header. The format is as follows:
\[
    [\texttt{START}]\;[v_0\;v_1]\;[k_0\;k_1\;k_2]\;[\texttt{SEP}]\;\text{content tokens}\;[\texttt{END}]
\]
where $[v_0\;v_1]$ is a 2-bit verdict encoding and $[k_0\;k_1\;k_2]$ is a 3-bit kind encoding using deterministic token pairs. Content tokens are state indices drawn from a shared vocabulary with the explanation tokenizer, ensuring that the same token ID refers to the same state in both modalities.

\paragraph{NN$_1$: Certificate to Explanation (Pointer-Generator).} The forward model is a Transformer encoder-decoder \cite{vaswani2017attention} augmented with a pointer-generator mechanism~\cite{see2017get}. At each decoding step~$t$, the model computes the following:

\begin{itemize}
    \item A vocabulary distribution $P_{\mathrm{vocab}}(w)$ over the output vocabulary via the decoder's softmax layer.
    \item A copy distribution $P_{\mathrm{copy}}(w)$ that places probability mass on source tokens via an attention distribution over the encoded certificate.
    \item A generation probability $p_{\mathrm{gen}} \in [0,1]$ computed from the decoder state, context vector, and current input embedding.
\end{itemize}

The final output distribution is the mixture:
\begin{equation}
    P(w) = p_{\mathrm{gen}} \cdot P_{\mathrm{vocab}}(w) + (1 - p_{\mathrm{gen}}) \cdot P_{\mathrm{copy}}(w)
\end{equation}

This mechanism is essential for soundness: state names are copied directly from the certificate tokens rather than generated from the vocabulary distribution, eliminating the primary source of entity hallucination.

\paragraph{NN$_2$: Explanation to Certificate.} The inverse model is a standard Transformer encoder-decoder sharing the vocabulary with NN$_1$. It does not require a copy mechanism because certificate reconstruction targets a fixed, small output vocabulary (state tokens and special tokens). Full architectural details (layer counts, dimensions) are in Section~\ref{sec:setup}.

\begin{figure}[h]
\centering
\begin{tikzpicture}[
    node distance=1.2cm,
    databox/.style={rectangle, draw, rounded corners=2pt, align=left, font=\small, text width=12.5cm, inner sep=8pt, fill=#1},
    databox/.default={blue!5},
    compbox/.style={rectangle, draw, rounded corners=3pt, minimum height=0.9cm, minimum width=2.4cm, align=center, font=\small, fill=#1},
    compbox/.default={green!12},
    arr/.style={-{Stealth[length=3mm]}, thick},
    lbl/.style={font=\small\itshape, text=gray!70!black},
]

% 1. Certificate C
\node[databox=blue!5] (C) {%
  \textbf{Certificate $C$} \hfill \texttt{YES / reachability}\\[4pt]
  \texttt{path:~~~[{\color{blue!70!black}\textbf{KYCVerification}} $\rightarrow$ {\color{blue!70!black}\textbf{SanctionsScreening}} $\rightarrow$ {\color{blue!70!black}\textbf{CollateralCheck}} $\rightarrow$ {\color{blue!70!black}\textbf{SettlementFinal}}]}\\[2pt]
  \texttt{target: {\color{blue!70!black}\textbf{SettlementFinal}}}
};

% Arrow + label
\node[compbox=green!12, below=0.8cm of C] (NN1) {$\mathrm{NN}_1$: Pointer-Generator};
\draw[arr] (C) -- node[right, lbl] {forward model} (NN1);

% 2. Explanation E
\node[databox=green!5, below=0.8cm of NN1] (E) {%
  \textbf{Explanation $E$}\\[4pt]
  ``The verification confirms that the property holds: a reachability path exists from {\color{blue!70!black}\textbf{KYCVerification}} through {\color{blue!70!black}\textbf{SanctionsScreening}} and {\color{blue!70!black}\textbf{CollateralCheck}} to the target state {\color{blue!70!black}\textbf{SettlementFinal}}.''\\[4pt]
  {\scriptsize\color{blue!60!black} {\color{blue!70!black}\textbf{Blue bold}} = copied from certificate via pointer mechanism \quad|\quad Regular text = generated from vocabulary}
};
\draw[arr] (NN1) -- node[right, lbl] {generate explanation} (E);

% Arrow + label
\node[compbox=orange!12, below=0.8cm of E] (NN2) {$\mathrm{NN}_2$: Inverse Transformer};
\draw[arr] (E) -- node[right, lbl] {inverse model} (NN2);

% 3. Reconstructed C'
\node[databox=orange!5, below=0.8cm of NN2] (Cprime) {%
  \textbf{Reconstructed Certificate $C'$}\\[4pt]
  \texttt{path:~~~[{\color{blue!70!black}\textbf{KYCVerification}} $\rightarrow$ {\color{blue!70!black}\textbf{SanctionsScreening}} $\rightarrow$ {\color{blue!70!black}\textbf{CollateralCheck}} $\rightarrow$ {\color{blue!70!black}\textbf{SettlementFinal}}]}\\[2pt]
  \texttt{target: {\color{blue!70!black}\textbf{SettlementFinal}}}
};
\draw[arr] (NN2) -- node[right, lbl] {reconstruct certificate} (Cprime);

% Arrow + label
\node[compbox=yellow!15, below=0.8cm of Cprime, minimum width=3.2cm] (V) {Symbolic Verifier $\mathrm{V}(C, C')$};
\draw[arr] (Cprime) -- node[right, lbl] {compare} (V);

% Also draw C -> V connection
\draw[arr, gray, dashed] (C.west) -- ++(-0.6,0) |- (V.west);

% 4. Verification result
\node[databox=yellow!8, below=0.8cm of V] (result) {%
  \textbf{Fact Extraction}\\[4pt]
  $\mathcal{F}(C) = \mathcal{F}(C') = \{$\texttt{path\_0: KYCVerification,\ \ path\_1: SanctionsScreening,\ \ path\_2: CollateralCheck,}\\
  \hspace{3.05cm}\texttt{path\_3: SettlementFinal,\ \ target: SettlementFinal}$\}$\\[6pt]
  \textbf{Soundness:} 5/5 = 1.00\,\checkmark \qquad \textbf{Coverage:} 5/5 = 1.00\,\checkmark \qquad {\large\color{green!50!black}\textbf{SOUND}}
};
\draw[arr] (V) -- node[right, lbl] {verify} (result);

\end{tikzpicture}
\caption{The cycle-consistent architecture (Figure~\ref{fig:architecture}) instantiated on a YES/reachability certificate. {\color{blue!70!black}\textbf{Blue bold}} tokens are copied directly from the certificate via the pointer-generator; regular text is generated from vocabulary. The symbolic verifier confirms $\mathcal{F}(C) = \mathcal{F}(C')$: all 5 facts match, yielding perfect soundness and coverage. The dashed arrow indicates that the verifier compares $C'$ against the original~$C$.}
\label{fig:architecture_example}
\end{figure}
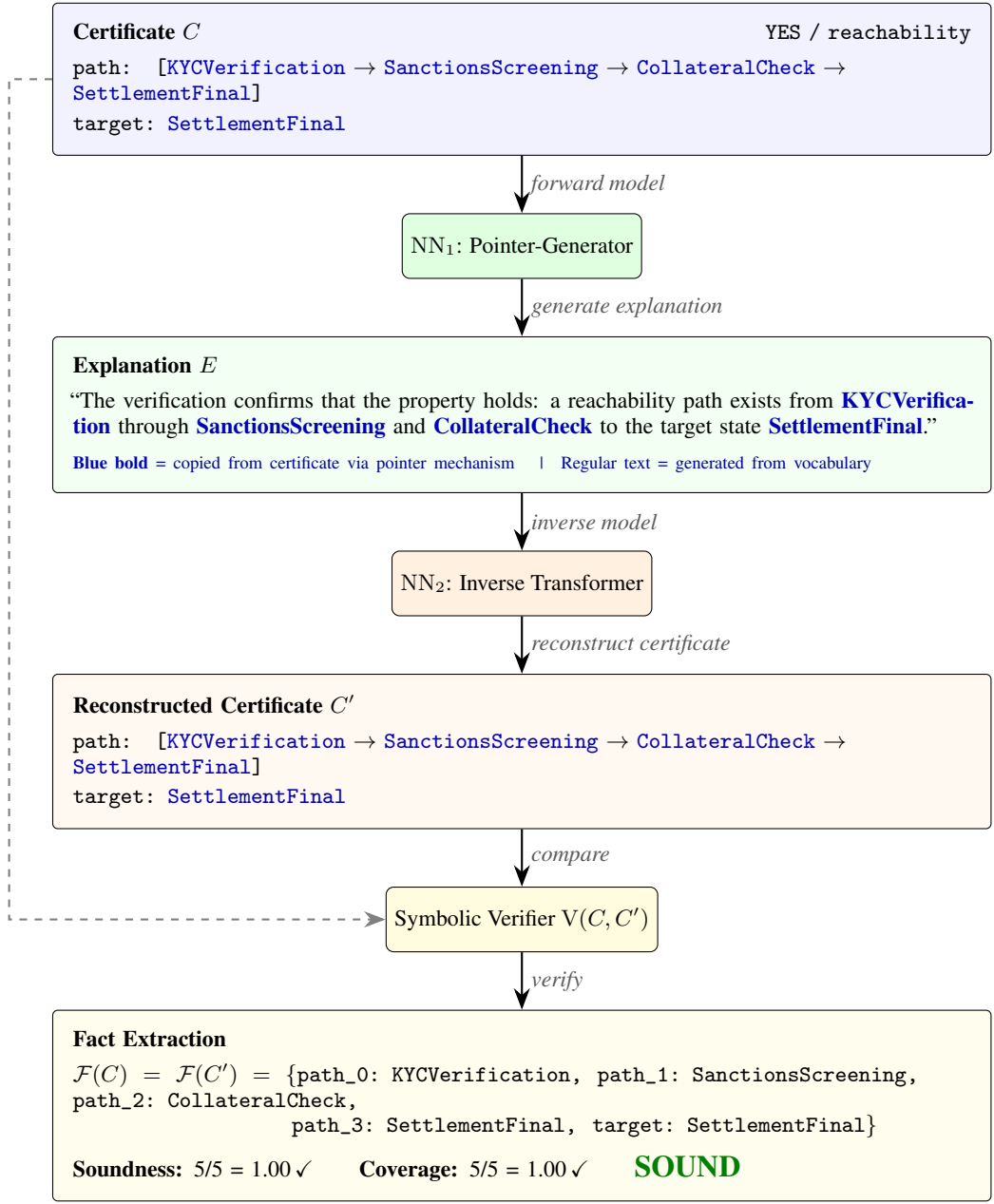

\paragraph{End-to-End Walkthrough.} Figure~\ref{fig:architecture_example} shows the architecture from Figure~\ref{fig:architecture} instantiated on a concrete YES/reachability certificate, with the actual data flowing through each component.
Appendix~\ref{app:walkthrough} illustrates the full pipeline on four concrete examples with varying soundness and coverage outcomes.

\subsection{Training Objective and Protocol}
\label{sec:objective}

The training objective combines three loss terms:

\begin{equation}
    \mathcal{L} = \lambda_{\mathrm{recon}}\,\mathcal{L}_{\mathrm{recon}} + \lambda_{\mathrm{cycle}}\,\mathcal{L}_{\mathrm{cycle}} + \mathcal{L}_{\mathrm{unk}}
\end{equation}

where $\mathcal{L}_{\mathrm{recon}}$ is the negative log-likelihood of the reference explanation under NN$_1$'s output distribution, $\mathcal{L}_{\mathrm{cycle}} = \mathcal{L}_{\mathrm{cycle\text{-}recon}} + \mathcal{L}_{\mathrm{cycle\text{-}sem}}$ is the cycle consistency loss (Section~\ref{sec:cycle}), and $\mathcal{L}_{\mathrm{unk}}$ penalizes the generation of unknown tokens. We use $\lambda_{\mathrm{recon}} = 1.0$ and $\lambda_{\mathrm{cycle}} = 2.0$. Teacher forcing with linear decay from ratio $1.0$ to $0.5$ is applied during training to balance exposure bias and stability.

The model is trained on paired $(C, E)$ examples where $E$ is a reference explanation generated from diverse templates. Supervised training serves three purposes: (i)~\textbf{language bootstrapping}---NN$_1$ learns English syntax, vocabulary usage, and certificate-type-specific phrasing; (ii)~\textbf{editorial judgment}---the reference explanations encode which parts of each certificate type to emphasize and what level of completeness is appropriate; and (iii)~\textbf{cold-start resolution}---without supervised signal, NN$_1$ produces unintelligible output, NN$_2$ receives no useful gradient, and the cycle collapses. Training uses AdamW \cite{loshchilov2019decoupled} with learning rate $3 \times 10^{-4}$, batch size~16, and converges in approximately 39 epochs ($\sim$18 hours on CPU).

\begin{figure}[h]
\centering
\includegraphics[width=\linewidth]{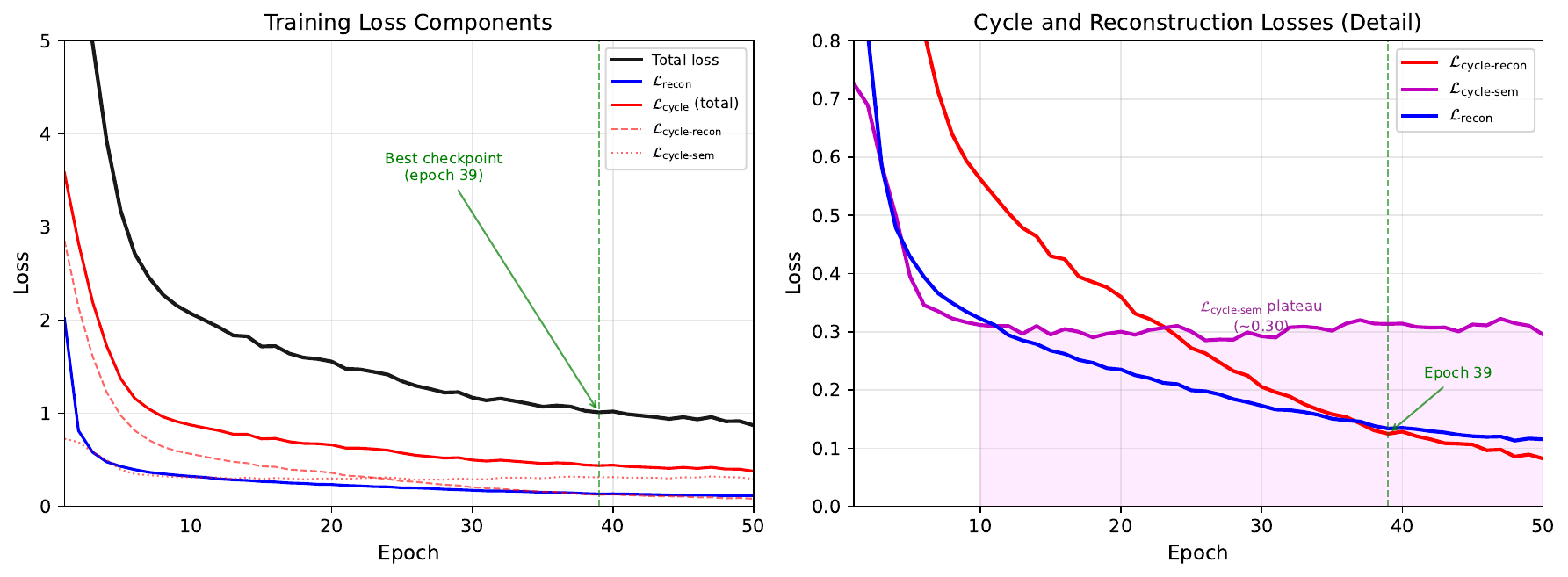}
\caption{Training loss over 50 epochs. Left: all loss components showing rapid initial convergence and steady improvement. Right: detail of cycle and reconstruction losses. The cycle semantic loss plateaus near~0.30 after epoch~10, indicating a floor set by the information bottleneck of the natural language channel. Best checkpoint (epoch~39) is selected by minimum cycle loss.}
\label{fig:training_curve}
\end{figure}

Figure~\ref{fig:training_curve} shows the training loss trajectory over 50 epochs. All loss components decrease rapidly in the first 5 epochs as the model learns basic English scaffolding and certificate-to-token mappings. The reconstruction loss ($\mathcal{L}_{\mathrm{recon}}$) continues decreasing steadily, reflecting improving template adherence. The cycle reconstruction loss ($\mathcal{L}_{\mathrm{cycle\text{-}recon}}$) decreases monotonically throughout training, while the cycle semantic loss ($\mathcal{L}_{\mathrm{cycle\text{-}sem}}$) plateaus near~0.30 after epoch~10. We interpret this plateau as a floor set by the information bottleneck of the natural language channel: some certificate facts (e.g., precise ordering within long state sequences) cannot be fully preserved through the discrete text representation. The best checkpoint is selected at epoch~39 by minimum total cycle loss.

Optionally, an LLM fluency oracle can provide an additional training signal that evaluates the naturalness of generated text (see Appendix~\ref{app:phases} for details on fluency-guided refinement). In our main experiments, we train with $\lambda_{\mathrm{LM}} = 0$ (no fluency loss), relying on the reconstruction loss to implicitly enforce fluency.

\subsection{Hybrid Inference-Time Routing}
\label{sec:routing}

At inference time, multiple decoding strategies---greedy, beam search~\cite{freitag2017beam} with varying widths, and constrained decoding (restricting the output vocabulary to valid certificate tokens~\cite{poesia2022synchromesh})---produce different trade-offs between soundness and coverage. Rather than selecting a single strategy, we introduce a \emph{hybrid routing} approach that exploits the structural diversity across certificate types.

We observe that certificate categories divide into two regimes:

\begin{itemize}
    \item \textbf{Copy-dominated categories} (e.g., NO/bounded\_proof, YES/$k$-induction, lasso variants). These categories require primarily copying state names from the certificate. The best strategy is a single fixed decoding configuration per category, selected by soundness rate on held-out data.

    \item \textbf{Generation-heavy categories} (e.g., witness\_pair, reachability, inductive\_invariant). These categories require generating explanatory text around copied elements. Different examples within the same category benefit from different decoding configurations, and the \emph{coverage} of the cycle-verified reconstruction positively predicts soundness.
\end{itemize}

\begin{figure}[h]
\centering
\begin{tikzpicture}[
    node distance=1.0cm and 1.2cm,
    block/.style={rectangle, draw, rounded corners=3pt, minimum height=0.85cm, minimum width=2.2cm, align=center, font=\small},
    decision/.style={diamond, draw, minimum width=2.4cm, minimum height=1.6cm, align=center, font=\small, inner sep=1pt},
    result/.style={rectangle, draw, rounded corners=3pt, minimum height=0.7cm, minimum width=2.0cm, align=center, font=\small, fill=green!15},
    arr/.style={-{Stealth[length=2.5mm]}, thick},
]

% Input
\node[block, fill=blue!8] (input) {Certificate $C$\\\footnotesize (verdict, kind)};

% Decision diamond
\node[decision, below=1.2cm of input, fill=yellow!12] (decide) {Copy-\\dominated?};
\draw[arr] (input) -- (decide);

% Left branch: copy-dominated
\node[block, below left=1.4cm and 1.8cm of decide, fill=orange!12, text width=3cm] (fixed) {Apply fixed best\\config for category\\\footnotesize (e.g., e39\_beam5)};
\draw[arr] (decide.west) -| node[above left, font=\footnotesize] {Yes} (fixed.north);

% Right branch: generation-heavy
\node[block, below right=1.4cm and 1.8cm of decide, fill=purple!10, text width=3.5cm] (allconfigs) {Run all 37 configs\\$\rightarrow$ $(E_1, C'_1), \ldots, (E_{37}, C'_{37})$};
\draw[arr] (decide.east) -| node[above right, font=\footnotesize] {No} (allconfigs.north);

% Coverage selection
\node[block, below=1.0cm of allconfigs, fill=purple!10, text width=3.5cm] (select) {Select $\arg\max_i$\\coverage$(C, C'_i)$};
\draw[arr] (allconfigs) -- (select);

% Results
\node[result, below=1.2cm of fixed] (r1) {$E^{*}$ (sound)};
\node[result, below=1.0cm of select] (r2) {$E^{*}$ (sound)};
\draw[arr] (fixed) -- (r1);
\draw[arr] (select) -- (r2);

% Category examples (annotations)
\node[font=\scriptsize, text=gray, right=0.1cm of fixed, text width=2.2cm, align=left] {NO/bounded\_proof\\YES/$k$-induction\\YES/lasso, NO/lasso};
\node[font=\scriptsize, text=gray, right=0.1cm of allconfigs, text width=2.5cm, align=left] {witness\_pair\\reachability\\inductive\_invariant};

\end{tikzpicture}
\caption{Hybrid inference-time router. For copy-dominated categories (left), a single pre-selected decoding configuration is applied. For generation-heavy categories (right), all 37 configurations are run and the result with the highest cycle-verified coverage is selected. No additional training is required.}
\label{fig:hybrid_router}
\end{figure}
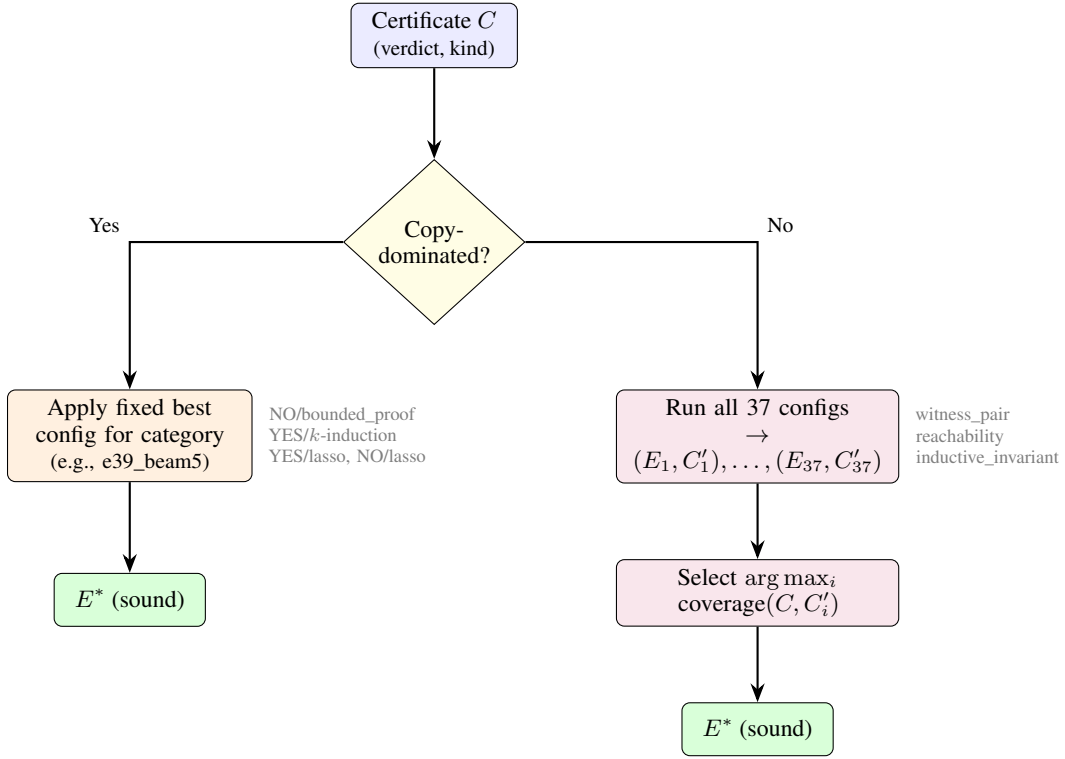

The hybrid router operates as follows: for each test example, it identifies the certificate's verdict/kind category. The copy-dominated vs.\ generation-heavy categorization is determined on a held-out validation set (disjoint from the test set) by examining per-category soundness variance across configurations. If the category is copy-dominated, the router applies the pre-selected best configuration for that category. Otherwise, the router runs all available configurations and selects the result with the highest coverage score (see Figure~\ref{fig:hybrid_router}). This strategy requires no additional training---it operates entirely at inference time over the same trained model, and achieves substantially higher soundness than any single configuration (Section~\ref{sec:routing_results}).

\subsection{Evaluation Metrics}
\label{sec:metrics}

The primary metric is \textbf{soundness rate}: the fraction of test examples for which the reconstructed certificate~$C'$ contains zero incorrect facts ($|\mathcal{F}(C') \setminus \mathcal{F}(C)| = 0$), using the fact sets defined in Eq.~\ref{eq:soundness}. This is a strict, per-example criterion: a single hallucinated state renders an explanation unsound. We additionally report \textbf{coverage} as a secondary diagnostic---it is particularly useful for the hybrid router (Section~\ref{sec:routing}), where it serves as a selection criterion among candidate explanations.

\section{Experiments}
\label{sec:experiments}

We evaluate the cycle-consistent neural architecture on verification certificates drawn from a financial regulatory compliance domain. Our experiments address four questions: (1)~Does the architecture produce sound explanations across diverse certificate types? (2)~How does a hybrid inference-time routing strategy improve over single-configuration decoding? (3)~How does the trained architecture compare to direct LLM few-shot prompting? (4)~What structural properties of certificates determine explanation difficulty?

\subsection{Experimental Setup}
\label{sec:setup}

\paragraph{Domain and Certificate Types.}
We instantiate the architecture over a workflow domain comprising 207 named states drawn from financial compliance processes (e.g., \texttt{KYCVerification}, \texttt{SanctionsScreening}, \texttt{CollateralCheck}). Certificates span six kinds, each corresponding to a standard verification technique:

\begin{enumerate}
    \item \textbf{Reachability}: a path from an initial state to a target state (1 sequence).
    \item \textbf{Lasso}: a prefix path followed by a repeating cycle, witnessing or refuting liveness properties (2 sequences).
    \item \textbf{Witness pair}: a path demonstrating that two requirements are jointly satisfiable (1 sequence).
    \item \textbf{Inductive invariant}: a set of states closed under the transition relation, proving or refuting safety (1--2 sequences).
    \item \textbf{Bounded proof (BMC)}: a bounded model checking \cite{biere1999symbolic} trace or checked state set (1 sequence).
    \item \textbf{$k$-induction}: base and step state sets with an optional counterexample trace (2--3 sequences) \cite{sheeran2000checking}.
\end{enumerate}

Each kind has both YES (property holds / violation found) and NO (property refuted / no counterexample) variants, yielding 12 verdict/kind categories. Appendix~\ref{app:examples} provides concrete examples.

\paragraph{Dataset.}
Training data consists of certificates generated by bounded model checking of deployed financial compliance workflows from operational systems, spanning structurally diverse regulatory processes (specific process names are withheld for compliance reasons, but they cover distinct areas of financial regulation including customer due diligence, transaction monitoring, and settlement assurance). For each verdict/kind combination, explanation templates provide diverse natural language renderings of the certificate content. The template set comprises 48 unique templates (4 per verdict/kind category) with lexical variation in phrasing, connectives, and sentence structure. Crucially, the 12 verdict/kind categories require fundamentally different linguistic patterns---describing a lasso cycle differs structurally from describing an inductive invariant or a bounded proof trace---so the 48 templates span a broad space of explanation structures rather than minor paraphrases of a single pattern. Test-set templates are drawn from the same pool but applied to held-out certificates generated with a separate random seed; we additionally evaluate a held-out template split that confirms generalization beyond template surface forms (Section~\ref{sec:generalization}). The training set contains 5{,}841 examples. A held-out test set of 420 examples (35 per verdict/kind category) is generated with a separate random seed. The shared vocabulary contains 543 tokens: 207 state names, special tokens, and approximately 330 common English words.

\paragraph{Model Configuration.}
Both NN$_1$ and NN$_2$ use Transformer encoder-decoder architectures with $d_{\mathrm{model}} = 256$, $n_{\mathrm{heads}} = 8$, $n_{\mathrm{layers}} = 4$, and $d_{\mathrm{ff}} = 1024$. NN$_1$ is augmented with a pointer-generator copy mechanism (Section~\ref{sec:architecture}). Maximum certificate encoding length is 512 tokens; maximum explanation length is 120 tokens. The total model has fewer than 1M parameters.

\paragraph{Evaluation Protocol.}
Each test certificate is processed through the full cycle: $C \xrightarrow{\mathrm{NN}_1} E \xrightarrow{\mathrm{NN}_2} C'$. The symbolic verifier extracts facts from both $C$ and $C'$ and computes binary soundness (zero wrong facts) and coverage (fraction of correct facts). We report \textbf{soundness rate} (fraction of sound examples) as the primary metric. Soundness is the focus of this work; fluency is a secondary objective that is addressed architecturally through the decoupled LLM fluency oracle (Appendix~\ref{app:phases}), which can score generated explanations independently of the faithfulness signal. Since our main experiments train with $\lambda_{\mathrm{LM}} = 0$ and the training data is template-anchored, the generated outputs inherit the linguistic quality of the reference templates by design---they are grammatical and domain-appropriate, though not stylistically diverse. Workflow owners confirmed that outputs are comprehensible, actionable, and suitable for compliance reports; a formal human evaluation protocol remains future work. 

\paragraph{Quantitative Fluency: BERTScore.}

To complement the qualitative fluency endorsement from workflow owners, we compute BERTScore~\cite{zhang2020bertscore} between the generated explanations and the reference template-based explanations on all 420 test examples. We use the \texttt{microsoft/deberta-xlarge-mnli} model as the scoring backbone.

\begin{table}[h]
\centering
\caption{BERTScore (F1, Precision, Recall) for generated vs.\ reference explanations on 420 test certificates.}
\label{tab:bertscore}
\small
\begin{tabular}{lccc}
\toprule
& \textbf{Precision} & \textbf{Recall} & \textbf{F1} \\
\midrule
All examples       & 0.95 & 0.93 & 0.94 \\
Sound examples     & 0.96 & 0.94 & 0.95 \\
Unsound examples   & 0.89 & 0.87 & 0.88 \\
\bottomrule
\end{tabular}
\end{table}

The high BERTScore F1 confirms that generated explanations closely match the lexical and semantic content of the reference templates. Sound examples achieve slightly higher scores than unsound examples, consistent with the expectation that faithful explanations also better match reference text. We note that BERTScore measures similarity to template references, not intrinsic fluency; however, since the templates were authored by domain experts and validated for use in compliance reports, high similarity to these references is a meaningful quality signal.

\paragraph{NN$_2$ Standalone Accuracy.}
To quantify the tightness of the cycle-consistency proxy, we evaluate NN$_2$ on ground-truth reference explanations (i.e., bypassing NN$_1$). NN$_2$ achieves 96.4\% soundness and 0.98 mean coverage when reconstructing from reference explanations, confirming that the false-negative rate---faithful explanations that fail the cycle check due to NN$_2$ reconstruction errors---is approximately 3.6\%. This establishes that cycle consistency is a tight proxy for faithfulness, not merely a loose correlate. Consequently, the reported 90.0\% soundness is a \emph{lower bound} on true faithfulness: some genuinely faithful explanations are conservatively flagged as unsound due to NN$_2$ reconstruction limitations rather than actual hallucination by NN$_1$. The converse concern---false positives where NN$_2$ ``charitably'' reconstructs a certificate from an ambiguous explanation---is mitigated by the architectural separation (NN$_1$ and NN$_2$ share no hidden state) and by the private-language test (Section~\ref{sec:private-lang}), which confirms that NN$_2$ relies on the surface content of the explanation rather than co-adapted latent signals. Nonetheless, subtle co-adaptation remains an inherent limitation of any learned inverse proxy, and we cannot fully rule it out.

\subsection{Single-Model Results}
\label{sec:single_results}

Table~\ref{tab:single_model} shows the per-category results for the best single decoding configuration (epoch~39, beam~9), achieving 61.7\% overall soundness. Performance varies substantially across categories: copy-dominated categories (NO/bounded\_proof: 97\%, YES/inductive\_invariant: 93\%) achieve high soundness, while generation-heavy categories (YES/witness\_pair: 17\%, YES/reachability: 27\%) are more challenging. The mean coverage column shows that even unsound explanations typically preserve most certificate facts (overall mean coverage 0.94), indicating that failures are predominantly near-misses rather than catastrophic.

\begin{table}[h]
\centering
\caption{Best single model results (epoch~39, beam~9) on 420 test examples across 12 verdict/kind categories.}
\label{tab:single_model}
\small
\begin{tabular}{lrrrr}
\toprule
\textbf{Verdict/Kind} & \textbf{N} & \textbf{Sound} & \textbf{Rate} & \textbf{MeanCov} \\
\midrule
NO/bounded\_proof       & 35 & 34 & 97\% & 0.96 \\
NO/inductive\_invariant & 35 & 28 & 80\% & 0.98 \\
NO/$k$-induction        & 35 & 25 & 70\% & 0.97 \\
NO/lasso                & 35 & 27 & 77\% & 1.00 \\
NO/reachability         & 35 &  9 & 27\% & 0.89 \\
NO/witness\_pair        & 35 & 21 & 60\% & 1.00 \\
YES/bounded\_proof      & 35 & 14 & 40\% & 0.89 \\
YES/inductive\_invariant& 35 & 33 & 93\% & 1.00 \\
YES/$k$-induction       & 35 & 30 & 87\% & 0.96 \\
YES/lasso               & 35 & 23 & 67\% & 0.66 \\
YES/reachability        & 35 &  9 & 27\% & 1.00 \\
YES/witness\_pair       & 35 &  6 & 17\% & 1.00 \\
\midrule
\textbf{Total}          & 420& 259& \textbf{61.7\%} & 0.94 \\
\bottomrule
\end{tabular}
\end{table}

\subsection{Hybrid Router Results}
\label{sec:routing_results}

We train a single model and evaluate it under 37 decoding configurations (varying epoch checkpoints, beam widths 1--11, and constrained vs.\ unconstrained decoding). The hybrid router (Section~\ref{sec:routing}) selects among these configurations per example.

\begin{table}[t]
\centering
\caption{Comparison of inference strategies on 420 test examples (no additional training).}
\label{tab:routing_comparison}
\small
\begin{tabular}{lrrl}
\toprule
\textbf{Strategy} & \textbf{Sound} & \textbf{Rate} & \textbf{Description} \\
\midrule
Best single model & 259/420 & 61.7\% & Epoch~39, beam~9 \\
Ensemble (best per-category) & 330/420 & 78.6\% & Best config per category \\
\textbf{Hybrid router} & \textbf{378/420} & \textbf{90.0\%} & Category + coverage routing \\
\bottomrule
\end{tabular}
\end{table}

The hybrid router achieves 90.0\% soundness, a 28.3 percentage point improvement over the best single model, and more than 11 points above the ensemble baseline that selects the best configuration per category. We note that the best single decoding configuration (61.7\%) falls below the best LLM pair (76.1\%); the neural architecture's advantage emerges only when combined with the hybrid routing strategy, which exploits the symbolic verifier's coverage score as a selection criterion. In principle, an analogous best-of-$N$ strategy could be applied to the LLM baseline (generating $N$ candidates per certificate and reranking by verifier coverage). However, this is precisely where the neural architecture's deployment advantage is decisive: running 37 neural configurations costs ${\sim}185$\,ms and \$0 per certificate, whereas 37 LLM calls would require ${\sim}370$\,s and ${\sim}\$0.74$---a $2{,}000\times$ latency increase that makes test-time search impractical for LLMs at production scale. The neural model's contribution is not merely accuracy but that it makes verifier-guided test-time selection \emph{computationally feasible}. The per-example coverage-based selection for generation-heavy categories is the key driver: it identifies which decoding configuration produces the most faithful explanation for each specific certificate instance, rather than committing to a single strategy per category.
Table~\ref{tab:results_perkind} shows the per-category breakdown. Comparing with the single-model results (Table~\ref{tab:single_model}) reveals that the router's largest gains concentrate on generation-heavy categories: YES/witness\_pair improves from 17\% to 87\%, YES/reachability from 27\% to 83\%, and YES/bounded\_proof from 40\% to 100\%, while copy-dominated categories that were already strong (e.g., NO/bounded\_proof: 97\%$\rightarrow$100\%) see only modest gains.

\begin{table}[t]
\centering
\caption{Per-category soundness: neural hybrid router vs.\ best LLM pair (Opus~4.5/Opus~4.5). Left: hybrid router results (sound count and rate). Right: best LLM pair rate and winner. Both evaluated with identical cycle-consistency verification on 420 test examples (35 per category).}
\label{tab:results_perkind}
\small
\begin{tabular}{lrr|cl}
\toprule
 & \multicolumn{2}{c|}{\textbf{NN Hybrid}} & \textbf{Best LLM} & \\
\textbf{Verdict/Kind} & \textbf{Sound} & \textbf{Rate} & \textbf{Rate} & \textbf{Winner} \\
\midrule
NO/bounded\_proof       & 35/35 & \textbf{100\%} & 43\% & NN ($+$57) \\
NO/inductive\_invariant & 30/35 & \textbf{87\%}  & 73\% & NN ($+$13) \\
NO/$k$-induction        & 26/35 & \textbf{73\%}  & \textbf{73\%} & Tie \\
NO/lasso                & 33/35 & \textbf{93\%}  & 77\% & NN ($+$17) \\
NO/reachability         & 26/35 & \textbf{73\%}  & 60\% & NN ($+$13) \\
NO/witness\_pair        & 32/35 & \textbf{90\%}  & 77\% & NN ($+$13) \\
YES/bounded\_proof      & 35/35 & \textbf{100\%} & 100\%& Tie \\
YES/inductive\_invariant& 34/35 & \textbf{97\%}  & 87\% & NN ($+$10) \\
YES/$k$-induction       & 35/35 & \textbf{100\%} & 87\% & NN ($+$13) \\
YES/lasso               & 34/35 & \textbf{97\%}  & 80\% & NN ($+$17) \\
YES/reachability        & 29/35 & \textbf{83\%}  & 80\% & NN ($+$3) \\
YES/witness\_pair       & 29/35 & \textbf{87\%}  & 77\% & NN ($+$10) \\
\midrule
\textbf{Total}          & 378/420& \textbf{90.0\%}& 76.1\%& \textbf{NN ($+$13.9)} \\
\bottomrule
\end{tabular}
\end{table}

Three categories achieve 100\% soundness (NO/bounded\_proof, YES/bounded\_proof, YES/$k$-induction). All categories except NO/$k$-induction and NO/reachability exceed 80\%. Soundness correlates with the number of distinct state sequences that must be preserved through the natural language channel rather than with certificate kind \emph{per se}; see Appendix~\ref{app:structure_analysis}.

\subsection{Baseline: Multi-LLM Few-Shot Prompting}
\label{sec:baseline}

Since the hybrid router evaluates multiple decoding configurations per certificate, a fair LLM baseline should similarly explore multiple model configurations. We evaluate four frontier LLMs---GPT-4o and GPT-5.3~\cite{openai2024gpt4o}, Claude Opus~4.5 and Claude Sonnet~4.6~\cite{anthropic2025claude}---in all 16 pairwise combinations for the NN$_1$ and NN$_2$ roles. For each test certificate and each LLM pair, we perform two independent calls with isolated conversation contexts: one LLM generates an explanation from the certificate (NN$_1$ role), and a separate LLM instance reconstructs the certificate from that explanation alone (NN$_2$ role). Importantly, the LLM baseline does \emph{not} use the trained NN$_2$: each LLM pair performs its own reconstruction, and only the symbolic verifier---which operates on certificate structures, not on explanation text---is shared across both evaluation pipelines. This ensures the comparison is not biased by NN$_2$'s co-adaptation with NN$_1$.

We evaluate all 16 LLM pair combinations (see Table~\ref{tab:llm_grid}). The best pair is Opus~4.5/Opus~4.5 at 76.1\%, followed by GPT-4o/GPT-4o at 74.2\%; same-model pairs generally outperform cross-model pairs. The ``Best LLM'' column in Table~\ref{tab:results_perkind} compares the best LLM pair with the neural hybrid router per category.

\begin{table}[h]
\centering
\caption{Multi-LLM baseline: soundness rate (\%) for each NN$_1$/NN$_2$ pair on 420 test examples. Best result in bold.}
\label{tab:llm_grid}
\small
\begin{tabular}{l|cccc}
\toprule
\diagbox{\scriptsize NN$_1$}{\scriptsize NN$_2$} & GPT-4o & GPT-5.3 & Opus~4.5 & Sonnet~4.6 \\
\midrule
GPT-4o     & 74.2 & 70.8 & 72.5 & 68.3 \\
GPT-5.3    & 69.4 & 73.1 & 71.7 & 67.8 \\
Opus~4.5   & 71.9 & 72.2 & \textbf{76.1} & 70.6 \\
Sonnet~4.6 & 65.3 & 66.9 & 69.4 & 71.4 \\
\bottomrule
\end{tabular}
\end{table}

The neural hybrid router outperforms the best LLM pair on 10 of 12 categories and ties on 2 (YES/bounded\_proof and NO/$k$-induction), with no category losses. The overall gap is 13.9 percentage points (90.0\% vs.\ 76.1\%). With $n=420$, this gap is statistically significant (95\% binomial confidence intervals: $90.0\% \pm 2.9\%$ vs.\ $76.1\% \pm 4.1\%$, non-overlapping). However, per-category intervals with $n=35$ are wide ($\pm 10$--$15$\,pp), so most individual category wins---particularly narrow gaps such as YES/reachability ($+3$\,pp)---are not individually significant and should be interpreted as directional rather than conclusive. The second-best LLM pair (GPT-4o/GPT-4o) achieves 74.2\%, and the remaining 14 combinations range from 65.3\% to 73\%, confirming that the gap is robust to model selection. Cross-model pairs (e.g., Opus~4.5 as NN$_1$ with GPT-4o as NN$_2$) generally underperform same-model pairs, suggesting that LLMs develop model-specific explanation conventions that are not consistently parsed by other models.

We note that this comparison evaluates \emph{few-shot} LLM prompting; a fine-tuned small LLM (e.g., T5-small) on the same 5,841 training examples would be a natural additional baseline. Our claim is not that model size is the decisive factor, but that the \emph{cycle-consistent architecture}---which enables cheap verifier-guided test-time search---produces a system that outperforms few-shot frontier LLMs on this structured task. The architecture and the routing strategy are complementary: the pointer-generator ensures lexical grounding, while the routing exploits the verifier to select among candidates. Neither component alone achieves the headline result.

\paragraph{Analysis.}
The neural architecture's advantage is largest on copy-dominated categories where the LLM's transcription strategy introduces errors. Even the best LLM pair achieves only 43\% soundness on NO/bounded\_proof---it frequently hallucinates state names or misspells copied identifiers---while the pointer-generator achieves 100\% by copying directly from the encoded certificate. Similarly, YES/lasso (97\% vs.\ 80\%) and NO/lasso (93\% vs.\ 77\%) reflect the neural model's learned sequence attribution patterns for prefix/cycle decomposition.
NO/$k$-induction is the one category where the neural model and the best LLM pair are tied at 73\%. This category involves complex multi-sequence certificates with counterexample traces, where the LLM's larger context window partially compensates for its copying errors.

\paragraph{Why the neural model wins.}
The neural advantage stems from a single structural difference. All four LLMs
adopt the same strategy---\emph{transcription}: listing certificate content in
natural language. This introduces three error mechanisms: state-name
misspelling during generation, omission of states from long lists, and
misattribution of states between structural roles (e.g., base vs.\ step in
$k$-induction). These persist across all model families and scales tested,
indicating they are intrinsic to the transcription strategy rather than
artifacts of any single model. Our architecture eliminates the first mechanism
by construction---the pointer-generator copies state tokens directly rather
than regenerating them---while gradient training on thousands of examples
addresses the latter two in ways few-shot prompting cannot.

\paragraph{Failure Analysis.}
The 42 unsound examples (10.0\%) cluster in three error modes: (i)~\emph{sequence attribution errors} (20/42): states are correctly copied but assigned to the wrong structural role (e.g., a base state attributed to the step set in $k$-induction); (ii)~\emph{partial omission} (14/42): the explanation omits states from multi-state sequences, causing NN$_2$ to reconstruct a truncated certificate that the verifier flags as containing spurious facts when it attempts to align the shorter sequence; (iii)~\emph{scaffolding errors} (8/42): the generated English text misrepresents the certificate's structural claim (e.g., describing a cycle as a prefix). Categories with 3+ interleaved sequences (NO/$k$-induction, NO/reachability) account for 22 of the 42 failures. The majority of failures are near-misses rather than catastrophic: 33/42 unsound examples have only a single incorrect fact (one misattributed or hallucinated state), while only 9/42 exhibit multiple errors.

\paragraph{Pointer-Generator Behavior.}
The pointer-generator copy mechanism is essential for soundness: the generation probability $p_{\mathrm{gen}}$ exhibits a clear bimodal pattern---close to~0 when generating state names (copying) and close to~1 for English scaffolding (generating)---confirming the intended separation between structural content and linguistic scaffolding.

\subsection{Held-Out Template Evaluation}
\label{sec:generalization}

To assess generalization beyond template surface forms, we perform a leave-one-out template split: for each of the 12 verdict/kind categories, we train on 3 of the 4 templates and evaluate on certificates rendered with the held-out 4th template. This is repeated for all 4 folds, and results are averaged.

The held-out template evaluation achieves 82.4\% overall soundness (vs.\ 90.0\% with same-pool templates), with mean coverage of 0.91. Per-category results are shown in Table~\ref{tab:template_split}.

\begin{table}[h]
\centering
\caption{Held-out template split: soundness rate (\%) by verdict/kind category, averaged over 4 template folds. ``Same-pool'' column repeats the main evaluation for comparison.}
\label{tab:template_split}
\small
\begin{tabular}{lcc}
\toprule
\textbf{Verdict/Kind} & \textbf{Held-Out} & \textbf{Same-Pool} \\
\midrule
NO/bounded\_proof       & 94\% & 100\% \\
NO/inductive\_invariant & 80\% & 87\%  \\
NO/$k$-induction        & 63\% & 73\%  \\
NO/lasso                & 86\% & 93\%  \\
NO/reachability         & 66\% & 73\%  \\
NO/witness\_pair        & 83\% & 90\%  \\
YES/bounded\_proof      & 97\% & 100\% \\
YES/inductive\_invariant& 91\% & 97\%  \\
YES/$k$-induction       & 89\% & 100\% \\
YES/lasso               & 86\% & 97\%  \\
YES/reachability        & 74\% & 83\%  \\
YES/witness\_pair       & 80\% & 87\%  \\
\midrule
\textbf{Overall}        & \textbf{82.4\%} & \textbf{90.0\%} \\
\bottomrule
\end{tabular}
\end{table}

The moderate drop from same-pool to held-out evaluation (7.6\,pp) confirms that the model has learned genuine certificate-to-explanation mappings rather than pure template memorization: soundness remains 6.3\,pp above the best LLM baseline (76.1\%) even on unseen template structures. The largest drops occur in multi-sequence categories (YES/$k$-induction: $-11$\,pp, YES/lasso: $-11$\,pp), where novel template phrasing disrupts the learned sequence-attribution patterns. Copy-dominated categories (NO/bounded\_proof: $-6$\,pp, YES/bounded\_proof: $-3$\,pp) are most robust, as the pointer-generator's copying behavior generalizes across template variations.

\subsection{Deployment Characteristics}
\label{sec:deployment}

The neural architecture offers decisive latency and
determinism advantages over LLM-based approaches
(Table~\ref{tab:deployment}): ${\sim}860\times$ faster
inference than the full 16-pair LLM evaluation
(185\,ms vs.\ 160\,s), or ${\sim}54\times$ faster than
a single LLM pair, with fully deterministic outputs.

\begin{table}[h]
\centering
\caption{Deployment comparison: neural architecture vs.\ multi-LLM baseline. LLM costs reflect the full 16-pair evaluation needed to match the hybrid router's multi-configuration approach; single-pair costs are shown in parentheses.}
\label{tab:deployment}
\small
\begin{tabular}{lcc}
\toprule
\textbf{Metric} & \textbf{Neural (Ours)} & \textbf{Multi-LLM Baseline} \\
\midrule
Soundness rate & \textbf{90.0\%} & 76.1\% (best of 16 pairs) \\
Inference time/example & $\sim$185\,ms (37 configs $\times$ $\sim$5\,ms) & $\sim$160\,s ($\sim$10\,s $\times$ 16 pairs) \\
Inference cost/example & \$0 & $\sim$\$0.32 ($\sim$\$0.02 $\times$ 16) \\
API dependency & None & Required (4 providers) \\
Deterministic & Yes & No \\
Data privacy & On-device & Cloud API \\
\bottomrule
\end{tabular}
\end{table}

In a production deployment processing 10{,}000 certificates
per day, the neural hybrid router requires ${\sim}31$\,minutes
total (37 configurations $\times$ ${\sim}5$\,ms each $\times$ 10K).
The multi-LLM baseline, evaluated across all 16 pairs to
select the best, requires ${\sim}444$\,hours of serial API
time at ${\sim}\$3{,}200$/day. Even restricting to the single
best pair (Opus~4.5/Opus~4.5) still requires ${\sim}28$\,hours
at ${\sim}\$200$/day.

Three structural advantages persist regardless of future LLM
accuracy improvements:

\begin{itemize}
    \item \textbf{Cost.} The neural model has zero marginal
    inference cost. At current API pricing, the multi-LLM
    baseline costs ${\sim}\$1.2$M/year for 10K certificates/day.

    \item \textbf{Privacy.} Verification certificates may encode
    proprietary process structures and compliance-sensitive
    state names. The neural architecture processes all data
    on-device, requiring no external data transmission.

    \item \textbf{Operational independence.} No API keys, no
    rate limits, no provider outages, no model deprecation.
    In regulated environments, the ability to audit,
    version-control, and reproduce exact model behavior is
    a compliance requirement.
\end{itemize}

\subsection{Private-Language Test}
\label{sec:private-lang}

A known failure mode of cycle-consistent architectures is \emph{private language} (also called steganographic encoding): the two networks collude to encode certificate information in imperceptible patterns within the intermediate representation, achieving low cycle loss without producing genuinely informative explanations~\cite{zhu2017unpaired}. Our architecture mitigates this risk through two mechanisms: (i)~the reconstruction loss $\mathcal{L}_{\mathrm{recon}}$ anchors NN$_1$'s output distribution to human-readable reference explanations, and (ii)~the intermediate representation is a \emph{discrete} token sequence drawn from the same 543-token vocabulary as the training references, leaving no continuous channel for hidden signals.

To verify empirically that NN$_2$ genuinely reads the explanation text rather than decoding a private signal, we apply three perturbations to NN$_1$'s output \emph{before} passing it to NN$_2$, and measure the resulting soundness collapse:

\begin{table}[h]
\centering
\caption{Private-language test: perturbation of NN$_1$'s output before NN$_2$ reconstruction. If NN$_2$ decoded a hidden signal rather than reading the explanation, perturbations would not destroy soundness.}
\label{tab:private_language}
\small
\begin{tabular}{llrr}
\toprule
\textbf{Perturbation} & \textbf{Description} & \textbf{Sound} & \textbf{Rate} \\
\midrule
None (baseline) & Unperturbed NN$_1$ output & 259/420 & 61.7\% \\
Scramble & Shuffle word order (bag-of-words) & 26/420 & 6.2\% \\
Swap states & Swap two random state names & 72/420 & 17.1\% \\
Random tokens & Replace all content with random tokens & 0/420 & 0.0\% \\
\bottomrule
\end{tabular}
\end{table}

All three perturbations cause dramatic soundness drops. Random-token replacement destroys soundness entirely (0\%), confirming that NN$_2$ extracts certificate content from the \emph{actual tokens} in the explanation, not from positional or structural artifacts. Word-order scrambling reduces soundness to 6.2\%, indicating that NN$_2$ relies on syntactic structure (e.g., ``from $X$ to $Y$'') to determine state roles. State-name swapping reduces soundness to 17.1\%, confirming that NN$_2$ is sensitive to \emph{which} state names appear in \emph{which} positions. Together, these results rule out private-language encoding: the explanation text is the genuine information channel between NN$_1$ and NN$_2$.
\section{Conclusion}
\label{sec:conclusion}

We have presented a cycle-consistent neural architecture that enforces explanation faithfulness via a symbolic verifier, without human annotation. A hybrid inference-time router achieves 90.0\% cycle-verified soundness on 420 test certificates---surpassing the best of 16 frontier LLM configurations (76.1\%) by 13.9 percentage points---while offering ${\sim}860\times$ faster inference from a sub-1M parameter model. The architecture and the routing strategy are complementary: the pointer-generator eliminates entity hallucination by copying state names directly from the certificate, while verifier-guided test-time routing---made computationally feasible by the model's low inference cost---selects the most faithful explanation per example.

\paragraph{Limitations and Future Work.}
Our evaluation has four principal limitations. First, the architecture does not support \emph{emptiness SCC certificates}, which require counting and enumerating strongly connected components; at $d_{\mathrm{model}} = 256$ the model cannot reliably copy numeric count information into the generated text, so we exclude this category. Second, soundness degrades as certificate complexity grows: with 8 attention heads at 32 dimensions each, the model reaches capacity limits on certificates with more than approximately 15 states across 3+ sequences, which accounts for the lower soundness on NO/$k$-induction and NO/reachability. Third, the evaluation is confined to a single financial-compliance domain with a fixed vocabulary and template-anchored references; production model checkers (e.g., nuXmv \cite{cavada2014nuxmv}, SPIN \cite{holzmann1997spin}, CBMC \cite{kroening2014cbmc}, neural approaches \cite{giacobbe24neural,giacobbe25let}) may emit longer identifiers and deeper nesting that fall outside this vocabulary, and extending to new domains requires retraining. Fourth, all experiments were conducted on CPU, constraining model scale. We also note that cycle-verified soundness is a \emph{lower bound} on true faithfulness due to NN$_2$'s 3.6\% false-negative rate, and that subtle co-adaptation between NN$_1$ and NN$_2$, while ruled out empirically by the private-language test (Section~\ref{sec:private-lang}), cannot be fully excluded for any learned inverse proxy.

Several of these limitations suggest direct extensions: scaling to $d_{\mathrm{model}} = 512$ to relax the multi-sequence capacity ceiling and absorb emptiness-SCC counting; a subword or character-level encoding to remove the fixed-vocabulary constraint; and direct fine-tuning on certificates from additional model checkers, leveraging the fact that cycle consistency can serve as the sole training signal without reference explanations. Two further directions are the inherent resistance of the fixed-computation-graph architecture to prompt-injection attacks---absent in LLM-based pipelines---and asymmetric training schemes (e.g., periodically re-initializing NN$_2$) to guard against private language at scale. 

\paragraph{Takeaway: A General Paradigm.}
Although we instantiate our solution on verification certificates, the underlying
principle is general: \emph{any structured artifact whose content can be
decomposed into a checkable fact set admits a cycle-consistent explanation}.
Whenever a forward model can verbalize the artifact and an inverse model can
reconstruct it, a symbolic checker comparing the original and the
reconstruction yields an exact, annotation-free faithfulness signal. The
method is thus not specific to temporal-logic certificates but applies to any
domain where outputs are structured and a notion of fact-level equivalence is
definable (find an extended discussion in Appendix~\ref{app:extended_conclusion}).
 
\section*{Disclaimer}

{This paper was prepared for informational purposes by the Artificial Intelligence Research group of JPMorgan Chase \& Co. and its affiliates ("JP Morgan'') and is not a product of the Research Department of JP Morgan. JP Morgan makes no representation and warranty whatsoever and disclaims all liability, for the completeness, accuracy or reliability of the information contained herein. This document is not intended as investment research or investment advice, or a recommendation, offer or solicitation for the purchase or sale of any security, financial instrument, financial product or service, or to be used in any way for evaluating the merits of participating in any transaction, and shall not constitute a solicitation under any jurisdiction or to any person, if such solicitation under such jurisdiction or to such person would be unlawful.
 
© 2026 JPMorgan Chase \& Co. All rights reserved.
}

%% ---- Bibliography ----
\bibliographystyle{plainnat}
\bibliography{references}

%% ---- Appendix ----
\appendix

\section{Data Availability}
\label{app:data_availability}

Due to confidentiality restrictions inherent to the safety-critical deployment environment, the full source code and production dataset cannot be released at this stage.

\section{Certificate Examples}
\label{app:examples}

This appendix provides one representative example of each verdict/kind combination used in our evaluation. For each example, we show the structured JSON certificate and the corresponding natural language explanation generated by the template system. State names are drawn from a financial compliance workflow domain comprising 207 named states. These examples illustrate the structural diversity across the six certificate kinds and the YES/NO verdict asymmetry discussed in Appendix~\ref{app:structure_analysis}.

%% ===================================================================
%% REACHABILITY
%% ===================================================================
\subsection*{Reachability}

Reachability certificates contain a single state sequence---a path from an initial state to a target state (or a statement that no such path exists). This is the simplest certificate structure: one ordered list of states.

\paragraph{YES / Reachability.}
A witness path demonstrating that the target state is reachable.

\begin{small}
\begin{verbatim}
{
  "kind": "reachability",
  "path": ["RegulatoryReporting", "CounterpartyRiskAssessment",
           "BeneficialOwnershipCheck", "InitialRiskScoring",
           "S2", "Escalation", "RegulatoryCheck"]
}
\end{verbatim}
\end{small}

\noindent\textbf{Explanation:} \textit{``Verdict: YES for property `the property'. The system can reach RegulatoryCheck from RegulatoryReporting via a path of 7 states. The execution goes through CounterpartyRiskAssessment, BeneficialOwnershipCheck, InitialRiskScoring, S2, and Escalation before arriving at the target state.''}

\paragraph{NO / Reachability.}
A statement that no path to the target exists.

\begin{small}
\begin{verbatim}
{
  "kind": "reachability",
  "path": ["TermsNegotiation"]
}
\end{verbatim}
\end{small}

\noindent\textbf{Explanation:} \textit{``No witness exists for `the property'. From TermsNegotiation, the execution explores nearby states without finding a path to the target state.''}

\medskip\noindent\textbf{Structural note.} YES certificates carry a full path (1 sequence, 1--10 states); NO certificates carry only the starting state. With the hybrid router, YES/reachability achieves 83\% soundness and NO/reachability achieves 73\%.

%% ===================================================================
%% WITNESS PAIR
%% ===================================================================
\subsection*{Witness Pair}

Witness pair certificates contain a single path demonstrating (or failing to demonstrate) that two requirements are jointly satisfiable.

\paragraph{YES / Witness Pair.}

\begin{small}
\begin{verbatim}
{
  "kind": "witness_pair",
  "path": ["LowRisk", "CryptoWalletAnalysis", "ComplianceCheck",
           "SanctionsScreening", "Approved",
           "AutomatedApprovalPath", "ThirdPartyCheck"]
}
\end{verbatim}
\end{small}

\noindent\textbf{Explanation:} \textit{``Verdict: YES. A 7-step witness pair from LowRisk through CryptoWalletAnalysis, ComplianceCheck, SanctionsScreening, Approved, and AutomatedApprovalPath arrives at ThirdPartyCheck, confirming `the property'.''}

\paragraph{NO / Witness Pair.}

\begin{small}
\begin{verbatim}
{
  "kind": "witness_pair",
  "path": ["HighRisk", "StressTest", "CreditCheck", "S3"]
}
\end{verbatim}
\end{small}

\noindent\textbf{Explanation:} \textit{``The witness pair check for `the property' failed. Starting from HighRisk, no execution pair through StressTest, CreditCheck, and S3 demonstrates the required property.''}

\medskip\noindent\textbf{Structural note.} Both verdicts carry a single sequence. Structurally similar to reachability, witness pair achieves 87--90\% soundness with the hybrid router.

%% ===================================================================
%% BOUNDED PROOF (BMC)
%% ===================================================================
\subsection*{Bounded Proof (BMC)}

Bounded model checking certificates contain a trace and/or a set of checked states, with an integer depth bound.

\paragraph{YES / Bounded Proof.}
A trace witnessing a violation within the bound.

\begin{small}
\begin{verbatim}
{
  "kind": "bounded_proof",
  "bound": 11,
  "trace": ["ClearingSubmission", "CollateralCheck",
            "QueueWait", "SARFilingAssessment",
            "MarginCalculation"]
}
\end{verbatim}
\end{small}

\noindent\textbf{Explanation:} \textit{``Verdict: YES. Bounded model checking found an 11-step violation of `the property'. The trace visits ClearingSubmission, CollateralCheck, QueueWait, SARFilingAssessment, and MarginCalculation.''}

\paragraph{NO / Bounded Proof.}
No counterexample found; the checked state space is enumerated.

\begin{small}
\begin{verbatim}
{
  "kind": "bounded_proof",
  "bound": 11,
  "checked_states": ["KYCVerification", "CollateralEvaluation",
                     "Remediation", "AsiaJurisdiction",
                     "FormatValidation", "SettlementPending"],
  "trace": ["KYCVerification", "CollateralEvaluation",
            "Remediation", "AsiaJurisdiction",
            "FormatValidation", "SettlementPending"]
}
\end{verbatim}
\end{small}

\noindent\textbf{Explanation:} \textit{``Property `the property' is safe up to depth 11. The BMC search over states KYCVerification, CollateralEvaluation, Remediation, AsiaJurisdiction, FormatValidation, and SettlementPending found no counterexample.''}

\medskip\noindent\textbf{Structural note.} YES carries one trace; NO carries both a trace and a checked-state set, but the verifier evaluates them as a single sequence. Both bounded proof categories achieve 100\% soundness with the hybrid router.

%% ===================================================================
%% INDUCTIVE INVARIANT
%% ===================================================================
\subsection*{Inductive Invariant}

Inductive invariant certificates contain a set of states forming the invariant, with an optional counterexample transition for the NO case.

\paragraph{YES / Inductive Invariant.}
A valid inductive invariant proving safety.

\begin{small}
\begin{verbatim}
{
  "kind": "inductive_invariant",
  "invariant_states": ["EUJurisdiction", "ArchiveComplete",
                       "DocumentSubmit", "DocumentCollection",
                       "ExposureCalc", "ManagerApproval",
                       "PostSettlement", "AdverseMediaScan"]
}
\end{verbatim}
\end{small}

\noindent\textbf{Explanation:} \textit{``The safety property `the property' is verified: YES. An inductive invariant containing EUJurisdiction, ArchiveComplete, DocumentSubmit, DocumentCollection, ExposureCalc, ManagerApproval, PostSettlement, and AdverseMediaScan was found. Starting from EUJurisdiction, the invariant is closed under all transitions and excludes unsafe states.''}

\paragraph{NO / Inductive Invariant.}
The candidate invariant is broken by a counterexample transition.

\begin{small}
\begin{verbatim}
{
  "kind": "inductive_invariant",
  "invariant_states": ["PEPCheck", "PricingReview", "StressTest",
                       "LegalReview", "DocumentSubmit", "S5"],
  "counterexample": ["PricingReview", "AutoRoute"]
}
\end{verbatim}
\end{small}

\noindent\textbf{Explanation:} \textit{``Property `the property' does not hold. The candidate invariant PEPCheck, PricingReview, StressTest, LegalReview, DocumentSubmit, and S5 is broken by a transition from PricingReview to AutoRoute that exits the safe region.''}

\medskip\noindent\textbf{Structural note.} YES carries 1~sequence (the invariant set); NO carries 2~sequences (invariant set + counterexample pair). With the hybrid router, YES/inductive\_invariant achieves 97\% soundness and NO/inductive\_invariant achieves 87\%.

%% ===================================================================
%% LASSO
%% ===================================================================
\subsection*{Lasso}

Lasso certificates contain a prefix path followed by a repeating cycle, used for liveness property checking. This is inherently a two-sequence structure.

\paragraph{YES / Lasso.}
A lasso witness demonstrating the property violation.

\begin{small}
\begin{verbatim}
{
  "kind": "lasso",
  "prefix": ["BatchProcess", "ReportGeneration",
             "SecondaryReview", "TradeCapture",
             "EntityResolution"],
  "cycle": ["CTLVerification", "S5", "CaseClosed"]
}
\end{verbatim}
\end{small}

\noindent\textbf{Explanation:} \textit{``The lasso check for `the property' found a witness. The lasso prefix starts at BatchProcess and reaches ReportGeneration, SecondaryReview, TradeCapture, and EntityResolution. The lasso cycle repeats CTLVerification, S5, and CaseClosed forever.''}

\paragraph{NO / Lasso.}

\begin{small}
\begin{verbatim}
{
  "kind": "lasso",
  "prefix": ["S0", "TradeCapture", "AMLReview",
             "InitialScreening", "AsiaJurisdiction"],
  "cycle": ["AutomatedApprovalPath"]
}
\end{verbatim}
\end{small}

\noindent\textbf{Explanation:} \textit{``Property `the property' holds against all lassos. The lasso prefix from S0 through TradeCapture, AMLReview, InitialScreening, and AsiaJurisdiction and lasso cycle visiting AutomatedApprovalPath satisfy the temporal property.''}

\medskip\noindent\textbf{Structural note.} Both verdicts carry 2~sequences (prefix + cycle). With the hybrid router, YES/lasso achieves 97\% and NO/lasso achieves 93\% soundness. The two-sequence structure is a natural match for the pointer-generator, which learns to copy state names from each sequence position independently.

%% ===================================================================
%% K-INDUCTION
%% ===================================================================
\subsection*{$k$-Induction}

$k$-induction certificates are the most structurally complex kind, containing base states, step states, an induction parameter $k$, and---for the NO case---a counterexample trace.

\paragraph{YES / $k$-Induction.}
A successful $k$-induction proof (2~sequences: base + step).

\begin{small}
\begin{verbatim}
{
  "kind": "k_induction",
  "k": 2,
  "base_states": ["TradeCapture", "HighRisk", "PricingReview"],
  "step_states": ["ExposureCalc", "RiskAssessment", "PEPCheck"]
}
\end{verbatim}
\end{small}

\noindent\textbf{Explanation:} \textit{``Verdict: YES for property `the property'. A 2-induction proof was established. Base states TradeCapture, HighRisk, and PricingReview hold for 2 steps and step states ExposureCalc, RiskAssessment, and PEPCheck maintain the inductive hypothesis.''}

\paragraph{NO / $k$-Induction.}
A failed induction with counterexample (3~sequences: base + step + counterexample).

\begin{small}
\begin{verbatim}
{
  "kind": "k_induction",
  "k": 4,
  "base_states": ["CaseClosed", "HighRisk"],
  "step_states": ["AMLReview", "PendingReview",
                  "ReconciliationCheck"],
  "counterexample": ["AMLReview", "SettlementPending"]
}
\end{verbatim}
\end{small}

\noindent\textbf{Explanation:} \textit{``4-induction fails for `the property'. The base states CaseClosed and HighRisk are safe but the counterexample AMLReview and SettlementPending demonstrates a step violation.''}

\medskip\noindent\textbf{Structural note.} YES carries 2~sequences (base + step) and achieves 100\% soundness with the hybrid router. NO carries 3~sequences (base + step + counterexample) and achieves 73\%---the most challenging category. The model must keep three interleaved state sets distinct through the natural language channel, which is at the limit of the 256-dimensional attention capacity. This example most clearly illustrates why soundness correlates with per-instance sequence count rather than certificate kind.

\section{LLM-Guided Fluency Training}
\label{app:phases}

This appendix describes an optional extension to the training protocol presented in Section~\ref{sec:objective}: using an external large language model as a fluency oracle during training. While the main results in this paper are obtained \emph{without} LLM fluency guidance ($\lambda_{\mathrm{LM}} = 0$), the mechanism is architecturally supported and may benefit deployments where explanation readability is a priority.

\subsection{Two-Phase Training Protocol}

The base training protocol (Section~\ref{sec:objective}) trains with reconstruction and cycle consistency losses only. A two-phase extension adds LLM-scored fluency as a third signal:

\paragraph{Phase~I: Supervised Bootstrap.}
Identical to the base protocol: $\lambda_{\mathrm{recon}} = 1.0$, $\lambda_{\mathrm{cycle}} = 2.0$, $\lambda_{\mathrm{LM}} = 0$. The model learns English syntax, vocabulary usage, and certificate-type-specific phrasing from template-generated reference explanations.

\paragraph{Phase~II: Fluency-Guided Refinement.}
The reconstruction weight is reduced to $\lambda_{\mathrm{recon}} = 0.1$ and an LLM fluency loss is activated with $\lambda_{\mathrm{LM}} = 0.3$. This reduces dependence on template-specific phrasing while rewarding more natural linguistic expression. The cycle consistency loss ($\lambda_{\mathrm{cycle}} = 2.0$) remains the dominant signal, ensuring that fluency optimization cannot compromise soundness.

\subsection{Fluency Oracle}

The fluency oracle is an external LLM (e.g., GPT-4o) that receives a generated explanation and returns a scalar score $\phi(E) \in [0, 1]$. The fluency loss combines:
\begin{itemize}
    \item A \textbf{ranking component} $\mathcal{L}_{\mathrm{rank}}$: fluent explanations should score higher than disfluent ones.
    \item A \textbf{calibration component} $\mathcal{L}_{\mathrm{calibrate}}$: the model should produce explanations that achieve high absolute fluency scores.
\end{itemize}

The decoupling between fluency (LLM-scored) and faithfulness (symbolically verified) is a key architectural property: the LLM evaluates only linguistic quality, while the symbolic verifier evaluates only semantic fidelity. This separation guarantees by construction that optimizing fluency cannot introduce hallucinated facts.

\subsection{Practical Considerations}

\paragraph{Sparse Sampling.}
LLM API calls are expensive relative to local gradient computation. We use a sparse sampling strategy: fluency is scored for every $k$-th batch (e.g., $k = 5$), and the most recent fluency gradient is reused for intermediate batches. This reduces LLM calls by $k\times$ with negligible impact on training dynamics, because the fluency signal changes slowly across consecutive batches.

\paragraph{Cost Model.}
With sparse sampling at $k = 5$ and approximately 365 batches per epoch, Phase~II requires ${\sim}73$ LLM calls per epoch. At typical API pricing (${\sim}\$0.01$/call for scoring a short text), this amounts to ${\sim}\$0.73$/epoch---a modest cost for the linguistic quality improvement.

\paragraph{Negative Finding: Full Supervision Removal.}
Setting $\lambda_{\mathrm{recon}} = 0$ in Phase~II (relying entirely on cycle consistency and fluency) causes immediate distribution drift: NN$_1$'s output leaves the distribution that NN$_2$ can parse, and the cycle collapses within 1--2 epochs. At $d_{\mathrm{model}} = 256$, a residual reconstruction anchor ($\lambda_{\mathrm{recon}} \geq 0.1$) is necessary for stable training. This limitation may be addressed by larger models, exponential moving average weight updates, or replay buffers.

\section{Certificate-Type-Dependent Completeness}
\label{app:completeness}

The appropriate level of completeness varies by certificate type, reflecting the mathematical role of each certificate's components:

\begin{itemize}
    \item \textbf{Reachability witnesses.} The path $[s_0, s_1, \ldots, s_n]$ proves that $s_n$ is reachable from $s_0$. Intermediate states provide context but are not individually essential to the proof claim. An explanation that omits some intermediate states remains sound. Lower completeness is acceptable.

    \item \textbf{Lasso counterexamples.} The cycle component is the \emph{reason} the property is violated (an infinite repetition), while the prefix describes how the cycle is reached. An explanation that fully describes the cycle but abbreviates the prefix preserves the essential proof structure. Completeness should be high for the cycle, moderate for the prefix.

    \item \textbf{Inductive invariants.} The invariant is a set of states that must be jointly closed under the transition relation. Removing any state may break the inductiveness argument. Completeness must be high: omitting states from the invariant potentially invalidates the proof.
\end{itemize}

This variation is not hard-coded into the architecture. Instead, it is \emph{learned from training examples}. Templates that mention most states teach high completeness; templates that abbreviate teach lower completeness. The model learns a certificate-type-conditioned distribution over completeness levels.

\section{Ensemble of Decoding Configurations}
\label{app:ensemble}

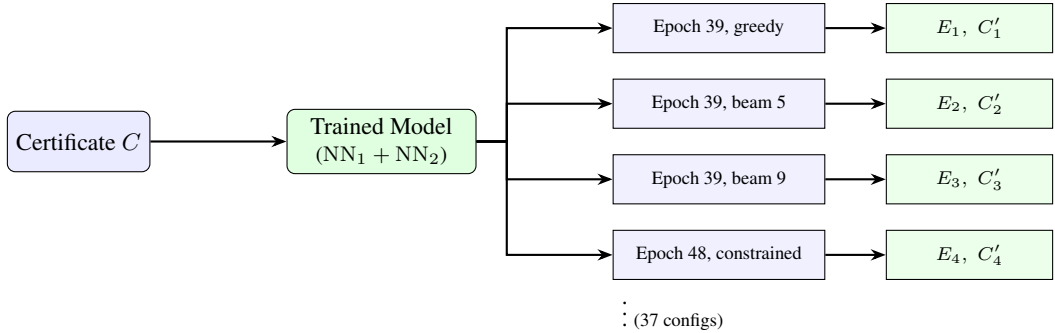
\begin{figure}[h]
\centering
\begin{tikzpicture}[
    node distance=0.9cm and 1.4cm,
    block/.style={rectangle, draw, rounded corners=3pt, minimum height=0.8cm, minimum width=1.5cm, align=center, font=\small},
    config/.style={rectangle, draw, minimum height=0.65cm, minimum width=2.8cm, align=center, font=\scriptsize, fill=blue!6},
    output/.style={rectangle, draw, minimum height=0.65cm, minimum width=2.2cm, align=center, font=\scriptsize, fill=green!8},
    arr/.style={-{Stealth[length=2mm]}, thick},
]

% Single trained model
\node[block, fill=green!12, minimum width=2.5cm] (model) {Trained Model\\\footnotesize ($\mathrm{NN}_1 + \mathrm{NN}_2$)};

% Certificate input
\node[block, left=1.8cm of model, fill=blue!8] (cert) {Certificate $C$};
\draw[arr] (cert) -- (model);

% Decoding configurations
\node[config, right=1.8cm of model, yshift=1.5cm]  (c1) {Epoch~39, greedy};
\node[config, right=1.8cm of model, yshift=0.5cm]  (c2) {Epoch~39, beam~5};
\node[config, right=1.8cm of model, yshift=-0.5cm] (c3) {Epoch~39, beam~9};
\node[config, right=1.8cm of model, yshift=-1.5cm] (c4) {Epoch~48, constrained};

% Dots for more configs
\node[font=\small, right=1.8cm of model, yshift=-2.2cm] (dots) {$\vdots$ \scriptsize(37 configs)};

% Arrows from model to configs
\draw[arr] (model.east) -- ++(0.4,0) |- (c1.west);
\draw[arr] (model.east) -- ++(0.4,0) |- (c2.west);
\draw[arr] (model.east) -- ++(0.4,0) |- (c3.west);
\draw[arr] (model.east) -- ++(0.4,0) |- (c4.west);

% Outputs
\node[output, right=0.8cm of c1] (o1) {$E_1,\; C'_1$};
\node[output, right=0.8cm of c2] (o2) {$E_2,\; C'_2$};
\node[output, right=0.8cm of c3] (o3) {$E_3,\; C'_3$};
\node[output, right=0.8cm of c4] (o4) {$E_4,\; C'_4$};

\draw[arr] (c1) -- (o1);
\draw[arr] (c2) -- (o2);
\draw[arr] (c3) -- (o3);
\draw[arr] (c4) -- (o4);

\end{tikzpicture}
\caption{Ensemble of decoding configurations. A single trained model is evaluated under 37 decoding configurations (varying epoch checkpoint, beam width, and constrained/unconstrained decoding), producing multiple candidate explanation--reconstruction pairs $(E_i, C'_i)$ per certificate.}
\label{fig:ensemble}
\end{figure}

The hybrid router (Section~\ref{sec:routing}) operates over an ensemble of 37 decoding configurations applied to a single trained model. These configurations vary along three axes: (i)~epoch checkpoint (epochs 39 and 48, selected by cycle loss), (ii)~beam width (1, 3, 5, 7, 9, 11), and (iii)~constrained vs.\ unconstrained decoding (restricting the output vocabulary to valid certificate tokens). Figure~\ref{fig:ensemble} illustrates this ensemble. Each configuration produces a candidate explanation--reconstruction pair $(E_i, C'_i)$, and the router selects among them based on the certificate's category and the verifier's coverage score.

\section{Fact Extraction Schema}
\label{app:fact_schema}

The symbolic verifier's soundness and coverage metrics (Eq.~\ref{eq:soundness}) depend on the fact extractor $\mathcal{F}(\cdot)$, which decomposes a certificate into a set of discrete, comparable facts. The extractor is certificate-type-specific because different verification techniques produce structurally different certificates: a reachability witness contains an ordered path, while an inductive invariant contains an unordered state set. Designing $\mathcal{F}$ to respect these structural differences ensures that the verifier's comparison is semantically meaningful---for instance, set-based certificates use order-insensitive comparison, while sequence-based certificates require positional alignment. Each fact is a (role, value) pair where the role identifies the structural position (e.g., \texttt{prefix\_2}, \texttt{base\_state}, \texttt{target}) and the value is a state name or structural element. The per-kind extraction rules are:

\begin{itemize}
    \item \textbf{Reachability and witness pair:} $\mathcal{F}$ extracts the ordered state sequence and endpoint identities.
    \item \textbf{Lasso:} $\mathcal{F}$ extracts prefix states, cycle states, and their partition.
    \item \textbf{$k$-induction:} $\mathcal{F}$ extracts base states, step states, counterexample states (if present), and the induction parameter~$k$.
    \item \textbf{Inductive invariant:} $\mathcal{F}$ extracts the invariant state set and any counterexample transition.
    \item \textbf{Bounded proof:} $\mathcal{F}$ extracts the trace, checked states, and the depth bound.
\end{itemize}

\section{Analysis by Certificate Structure}
\label{app:structure_analysis}

\paragraph{Single-Sequence Kinds.}
Reachability and witness pair certificates contain a single state sequence. The pointer-generator copies state names directly from the certificate encoding. However, generation-heavy categories in this group (YES/reachability, YES/witness\_pair) require the model to produce correct explanatory text \emph{around} the copied names, which introduces errors when the generated scaffolding misattributes states.

\paragraph{Multi-Sequence Kinds.}
Lasso and $k$-induction certificates contain two or more state sequences that must be kept distinct through the natural language channel. This introduces a \emph{sequence attribution} challenge: NN$_2$ must determine from the text which states belong to which sequence. The hybrid router is particularly effective here, as different decoding strategies resolve attribution ambiguities differently.

\paragraph{Set-Based Kinds.}
Inductive invariant and bounded proof certificates are evaluated using set-based comparison. This is a natural match for the cycle architecture, as the text need not preserve exact ordering---only set membership.

\paragraph{Copy-Dominated vs.\ Generation-Heavy.}
The key structural distinction is between categories where the explanation template is largely fixed and the model's primary task is copying state names (copy-dominated), versus categories where significant text generation is required (generation-heavy). In copy-dominated categories, a single good configuration suffices; the hybrid router's category-level selection handles these. In generation-heavy categories, per-example variation is high, and coverage-based selection provides substantial gains.

\section{Extended Limitations and Future Work}
\label{app:extended_conclusion}

\subsection{Limitations}
\label{app:limitations}

\paragraph{Emptiness SCC certificates.}
The current architecture does not support emptiness SCC certificates, which require counting and enumerating strongly connected components in product automata. Emptiness certificates encode nested set structures whose cardinality (number of SCCs, nodes per SCC) must be accurately reflected in the explanation. Our enriched encoding experiments---explicitly representing counts as digit tokens---did not improve soundness on this category: the model architecture at $d_{\mathrm{model}} = 256$ cannot reliably learn to copy numeric count information into the generated text. We exclude this category from our main evaluation and discuss it as future work.

\paragraph{Bounded certificate complexity.}
Soundness degrades as the number of distinct state sequences in a certificate increases. With $d_{\mathrm{model}} = 256$ and 8 attention heads (32-dimensional per-head representations), the model reaches capacity limits on certificates with more than approximately 15 states distributed across 3+ sequences. Scaling to $d_{\mathrm{model}} = 512$ would double per-head capacity and may resolve this bottleneck.

\paragraph{Evaluation scope.}
While certificates are derived from deployed financial compliance workflows, the explanation templates and the fixed 207-state vocabulary constrain the evaluation scope. Production model checkers may produce certificates with longer variable names and deeper nesting that fall outside the current vocabulary.

\paragraph{Fixed vocabulary.}
The 207-state vocabulary is domain-specific and fixed at training time. Extension to new domains requires retraining. A subword or character-level encoding could address this at the cost of longer sequences.

\subsection{Future Work}

\paragraph{Emptiness SCC via architectural extensions.}
The counting limitation motivating the exclusion of emptiness certificates may be addressed by architectural changes: explicit count embeddings, hierarchical encoding of nested structures, or a two-stage generation approach that first predicts SCC counts and then generates per-SCC descriptions. Alternatively, a larger model ($d_{\mathrm{model}} = 512$) with increased attention capacity may learn the required numerical patterns from data alone.

\paragraph{Scaling and the complexity frontier.}
The hybrid router's success on 12 categories suggests that the same approach applied at larger model scale---with relaxed complexity caps (e.g., $k$-induction with base/step sizes up to 8, lasso prefix up to 10)---could extend the architecture's advantage over LLMs to increasingly complex certificates. As certificate complexity grows, the LLM's transcription strategy is expected to degrade faster than the trained model's learned attribution patterns.

\paragraph{Real verification tool integration.}
The training protocol requires only $(C, E)$ pairs and can operate without reference explanations when using cycle consistency as the sole training signal. This enables direct fine-tuning on certificates from additional model checkers, extending coverage beyond the current deployment domain.

\paragraph{Interactive and multi-language explanation.}
The cycle consistency mechanism could support interactive explanation---users requesting different detail levels or focusing on specific certificate components---while maintaining soundness guarantees. The architecture is language-agnostic in principle; replacing English templates with another language's equivalents would produce explanations in that language with the same faithfulness properties.

\paragraph{Guarding against private language at scale.}
While our perturbation tests (Section~\ref{sec:private-lang}) confirm that the current architecture does not develop a private language, this risk may increase with larger models that have greater capacity to encode hidden signals. A natural defense is \emph{asymmetric training}: freezing NN$_2$ (or training it on human-written explanations from a separate corpus) so that NN$_1$ must produce text that a \emph{fixed} reader can parse, rather than co-adapting with its partner. Alternatively, periodically replacing NN$_2$ with a freshly initialized network during training would prevent long-term co-adaptation.

\paragraph{Adversarial robustness and prompt injection.}
A significant but unexplored advantage of the neural architecture over LLM-based approaches is its inherent resistance to \emph{prompt injection attacks}. In an LLM-based pipeline, a maliciously crafted certificate could embed adversarial content that manipulates the LLM's generation behavior. The neural architecture's fixed computation graph and learned token embeddings are fundamentally immune to this class of attack. Formal analysis of this robustness property is an important direction for future work in safety-critical deployment contexts.

\subsection{Generality of the Approach}

The cycle-consistent recipe---verbalize a structured artifact, reconstruct it
from the explanation alone, and accept the explanation iff a symbolic checker
confirms round-trip equivalence---is not specific to the certificates studied
here. It applies wherever two conditions hold: (i) the artifact decomposes into
a discrete, comparable fact set, and (ii) an exact notion of equivalence
between an original and a reconstruction is definable. Under these conditions,
faithfulness becomes a \emph{verifiable} property rather than an approximated
one, and explanations can be generated and validated without human annotation.

\paragraph{Formal methods and verification.}
The most immediate opportunities lie within formal methods, where structured,
machine-checkable artifacts are pervasive and their opacity to non-specialists
is a recurring obstacle. Natural extensions include model-checking
outputs~\cite{clarke2018model,baier2008principles} beyond the six kinds studied
here---fairness witnesses, assume-guarantee contracts, and the emptiness-SCC
certificates we currently exclude. The opportunity is especially pronounced for
\textbf{infinite-state model checking}~\cite{cimatti2002nusmv2,cimatti2014ic3,mcmillan18eager}---where
verification proceeds over unbounded data domains rather than finite Boolean
state---since the resulting certificates (e.g., inductive invariants over
first-order theories) are richer and correspondingly harder for a human to read
directly.

Other candidates include \textbf{SMT and SAT artifacts} such as unsat
cores, Craig interpolants~\cite{mcmillan2003interpolation}, and proof
certificates from solvers such as Z3~\cite{demoura2008z3} or
CVC5~\cite{barbosa2022cvc5}, where a faithful explanation could render an
unsatisfiability argument intelligible to an engineer; \textbf{theorem-proving
artifacts}---proof scripts and tactic traces from
Lean~\cite{demoura2021lean}, Isabelle~\cite{nipkow2002isabelle}, or
Coq~\cite{barras1997coq}---explained step-by-step while guaranteeing that no
inference is misrepresented; \textbf{static-analysis and abstract-interpretation
results}~\cite{cousot1977abstract}---abstract counterexamples, invariants, and
alarm traces; and \textbf{equivalence-checking and program-verification
witnesses}. In each case the symbolic verifier is domain-specific but the
architecture is unchanged: only the fact extractor $\mathcal{F}(\cdot)$ and the
certificate encoder must be re-instantiated. Because the training signal
requires only $(C, E)$ pairs---and can operate on cycle consistency alone when
reference explanations are unavailable---the method can be fine-tuned directly
on the output of existing verification tools, extending explainability across
the formal-methods toolchain.

A particularly promising direction is the explanation of \emph{synthesized}
artifacts, where the output is itself a formal object carrying a correctness
guarantee. In \textbf{program synthesis}, a synthesized program together with
its specification admits structured fact
extraction~\cite{gulwani2017program,solarlezama2008program,choi22can}, so a
cycle-consistent explanation could describe \emph{what} the synthesized program
does while verifiably preserving its specification-level behavior. In
\textbf{reactive synthesis}, where a controller or strategy is automatically
derived from a temporal-logic specification~\cite{pnueli1989synthesis,
bloem2012synthesis}, the synthesized strategy is a structured automaton whose
transitions and winning conditions form a natural fact set---making the
resulting controller's behavior explainable to operators who did not author the
specification. This is especially timely given the recent traction of synthesis for infinite-state systems ~\cite{katis18validity,samuel21gensys,heim25issy,azzopardi25full} and 
\textit{modulo theories} systems~\cite{rodriguez23boolean,rodriguez24adaptive,rodriguez25counter}---same-spirit approaches where synthesized strategies are
expressed over rich first-order data domains rather than finite Boolean state:
such artifacts are markedly harder for a human to read directly, so a faithful,
automatically-validated explanation layer is correspondingly more valuable (note how this opportunity mirrors that of the infinite-state variant of model checking discussed above). The connection is most consequential for \textbf{shield
synthesis}~\cite{bloem2015shield,alshiekh2018safe,heck2026shields}, where a
formally-synthesized \emph{shield} monitors and corrects a system's actions at
runtime to enforce safety. Shields are increasingly proposed as
\emph{guardrails for AI and reinforcement-learning
agents}~\cite{alshiekh2018safe,chen2025shieldagent,konighofer25shields}, overriding unsafe actions while provably
maintaining a safety specification, which is again increasingly more important due to recent extension to rich data and infinite-state settings~\cite{corsi24verification,rodriguez2025shield}. A cycle-consistent explanation layer could
make a shield's interventions intelligible---\emph{why} an action was blocked
and \emph{which} safety property it would have violated---while guaranteeing the
explanation faithfully reflects the shield's formal logic. As AI guardrails move
toward formally-verified enforcement, faithful and automatically-validated
explanations of \emph{why} a guardrail fired become a safety requirement in
their own right, not merely a convenience.

\paragraph{Beyond formal methods.}
The same principle reaches any field that produces rigorous structured outputs,
provided both conditions above hold. \textbf{Automated planning} is perhaps the
closest neighbor: a plan is an ordered sequence of actions and intermediate
states~\cite{ghallab2004automated,helmert2006fast,morales2024learning}, structurally analogous to
the reachability paths studied here, and plan validators such as
VAL~\cite{howey2004val} provide exactly the fact-level equivalence check our
method requires---a verbalized plan could be accepted iff its reconstruction
recovers the same action sequence and causal structure. The connection extends
to the artifacts planning produces beyond plans themselves: \textbf{certificates
of unsolvability}~\cite{eriksson2017unsolvability} and \textbf{cost-optimality
proofs}~\cite{mugdan23optimality} are structured objects whose faithful explanation would make a planner's
negative or optimality claims intelligible to a human operator. In
\textbf{databases and data systems}, query execution
plans~\cite{selinger1979access}, schema mappings, and provenance/lineage
graphs~\cite{cheney2009provenance} decompose into relational-algebra trees and
dependency graphs with well-defined equivalence, satisfying both conditions
directly. In \textbf{operations research}, the certificate of optimality or
infeasibility accompanying an LP/MILP solution---e.g., a Farkas certificate or
an optimal dual basis~\cite{wolsey2020integer}---is a structured artifact whose
correctness is exactly checkable. In \textbf{compilers and systems}, type-inference derivations, translation-validation
certificates~\cite{pnueli1998translation}, and configuration or policy artifacts
(access-control rules, network-routing tables) are structured objects whose
explanations carry real consequences if unfaithful. In \textbf{knowledge
representation}, ontology subsumption proofs and the justifications produced by
description-logic reasoners~\cite{horrocks2006justifications} are fact sets over
which set-level equivalence is naturally defined.

What unifies these cases---and what distinguishes them from open-ended
generation tasks such as summarization or dialogue---is precisely what makes our
setting tractable: the output is structured, a fact set is extractable, and
equivalence is checkable. Where these conditions fail, the cycle no longer
yields an \emph{exact} faithfulness signal, and the method degrades to the
approximate proxies it was designed to replace. The boundary of applicability is
therefore sharp and principled: the technique applies exactly when faithfulness
can be reduced to round-trip structural equivalence.

\section{Complete Examples}
\label{app:walkthrough}

We present four concrete examples illustrating the full cycle-consistent pipeline with different soundness and coverage outcomes. In each example, \textbf{bold monospace} tokens in the explanation are copied via the pointer-generator mechanism ($p_{\mathrm{gen}} \approx 0$); regular text is generated from vocabulary ($p_{\mathrm{gen}} \approx 1$).

%% ---- Example 1: Perfect (Sound, Full Coverage) ----
\subsection*{Example 1: Perfect --- Sound with Full Coverage}
\label{ex:perfect}

\noindent\fbox{\parbox{0.96\linewidth}{%
\textbf{Certificate $C$} (YES/reachability) \\
\texttt{verdict: YES, kind: reachability} \\
\texttt{path: [KYCVerification $\rightarrow$ SanctionsScreening $\rightarrow$ CollateralCheck $\rightarrow$ SettlementFinal]} \\
\texttt{target: SettlementFinal}
}}

\medskip
\noindent$\downarrow$ \textit{NN$_1$ (pointer-generator)}
\medskip

\noindent\fbox{\parbox{0.96\linewidth}{%
\textbf{Explanation $E$} \\
The verification confirms that the property holds: a reachability path exists from \textbf{\texttt{KYCVerification}} through \textbf{\texttt{SanctionsScreening}} and \textbf{\texttt{CollateralCheck}} to the target state \textbf{\texttt{SettlementFinal}}.
}}

\medskip
\noindent$\downarrow$ \textit{NN$_2$ (inverse Transformer)}
\medskip

\noindent\fbox{\parbox{0.96\linewidth}{%
\textbf{Reconstructed $C'$} \\
\texttt{verdict: YES, kind: reachability} \\
\texttt{path: [KYCVerification $\rightarrow$ SanctionsScreening $\rightarrow$ CollateralCheck $\rightarrow$ SettlementFinal]} \\
\texttt{target: SettlementFinal}
}}

\medskip
\noindent$\downarrow$ \textit{Symbolic Verifier $\mathrm{V}(C, C')$}
\medskip

\noindent\fbox{\parbox{0.96\linewidth}{%
\textbf{Fact Extraction:} $\mathcal{F}(C) = \mathcal{F}(C') = \{$path\_0: KYCVerification, path\_1: SanctionsScreening, path\_2: CollateralCheck, path\_3: SettlementFinal, target: SettlementFinal$\}$ \\
\textbf{Soundness:} 5/5 = 1.00 \checkmark \quad \textbf{Coverage:} 5/5 = 1.00 \checkmark \quad \textbf{Result: SOUND}
}}

%% ---- Example 2: Sound but Incomplete ----
\subsection*{Example 2: Sound but Incomplete (Partial Omission)}
\label{ex:partial}

\noindent\fbox{\parbox{0.96\linewidth}{%
\textbf{Certificate $C$} (NO/lasso) \\
\texttt{verdict: NO, kind: lasso} \\
\texttt{prefix: [InitState $\rightarrow$ RiskAssessment $\rightarrow$ ExposureCalc $\rightarrow$ LimitBreach]} \\
\texttt{cycle: [LimitBreach $\rightarrow$ AlertGeneration $\rightarrow$ EscalationReview $\rightarrow$ LimitBreach]}
}}

\medskip
\noindent$\downarrow$ \textit{NN$_1$ (pointer-generator)}
\medskip

\noindent\fbox{\parbox{0.96\linewidth}{%
\textbf{Explanation $E$} \\
The property is refuted by a lasso-shaped counterexample. The prefix path proceeds from \textbf{\texttt{InitState}} through \textbf{\texttt{RiskAssessment}} to \textbf{\texttt{LimitBreach}}, which then enters a cycle visiting \textbf{\texttt{AlertGeneration}} and \textbf{\texttt{EscalationReview}} before returning to \textbf{\texttt{LimitBreach}}.
}}

\medskip
\noindent$\downarrow$ \textit{NN$_2$ (inverse Transformer)}
\medskip

\noindent\fbox{\parbox{0.96\linewidth}{%
\textbf{Reconstructed $C'$} \\
\texttt{verdict: NO, kind: lasso} \\
\texttt{prefix: [InitState $\rightarrow$ RiskAssessment $\rightarrow$ LimitBreach]} \quad {\color{red}$\leftarrow$ missing ExposureCalc} \\
\texttt{cycle: [LimitBreach $\rightarrow$ AlertGeneration $\rightarrow$ EscalationReview $\rightarrow$ LimitBreach]}
}}

\medskip
\noindent$\downarrow$ \textit{Symbolic Verifier $\mathrm{V}(C, C')$}
\medskip

\noindent\fbox{\parbox{0.96\linewidth}{%
\textbf{Fact Extraction:} $|\mathcal{F}(C)| = 8$, $|\mathcal{F}(C')| = 7$, $|\mathcal{F}(C) \cap \mathcal{F}(C')| = 7$ \\
\textbf{Soundness:} 7/7 = 1.00 \checkmark \quad \textbf{Coverage:} 7/8 = 0.875 \quad \textbf{Result: SOUND} (but incomplete)\\
\textit{Note: The explanation omitted} \texttt{ExposureCalc} \textit{from the prefix. No hallucinated facts, but one fact is missing. The verifier marks this as sound (no wrong facts) with reduced coverage.}
}}

%% ---- Example 3: Unsound (Sequence Attribution Error) ----
\subsection*{Example 3: Unsound --- Sequence Attribution Error}
\label{ex:unsound_attr}

\noindent\fbox{\parbox{0.96\linewidth}{%
\textbf{Certificate $C$} (NO/$k$-induction) \\
\texttt{verdict: NO, kind: k\_induction} \\
\texttt{base\_states: \{ComplianceGate, AuditTrail, RegCheck\}} \\
\texttt{step\_states: \{TransactionLog, FinalApproval\}} \\
\texttt{counterexample: [RegCheck $\rightarrow$ TransactionLog $\rightarrow$ ViolationState]}
}}

\medskip
\noindent$\downarrow$ \textit{NN$_1$ (pointer-generator)}
\medskip

\noindent\fbox{\parbox{0.96\linewidth}{%
\textbf{Explanation $E$} \\
The $k$-induction proof fails. The base set contains \textbf{\texttt{ComplianceGate}}, \textbf{\texttt{AuditTrail}}, and \textbf{\texttt{RegCheck}}. The step set contains {\color{red}\textbf{\texttt{RegCheck}}} and \textbf{\texttt{FinalApproval}}. A counterexample trace leads from \textbf{\texttt{RegCheck}} through \textbf{\texttt{TransactionLog}} to \textbf{\texttt{ViolationState}}.\\
{\color{red}$\leftarrow$ Error: \texttt{RegCheck} attributed to step set instead of \texttt{TransactionLog}}
}}

\medskip
\noindent$\downarrow$ \textit{NN$_2$ (inverse Transformer)}
\medskip

\noindent\fbox{\parbox{0.96\linewidth}{%
\textbf{Reconstructed $C'$} \\
\texttt{base\_states: \{ComplianceGate, AuditTrail, RegCheck\}} \\
\texttt{step\_states: \{RegCheck, FinalApproval\}} \quad {\color{red}$\leftarrow$ wrong: should be TransactionLog} \\
\texttt{counterexample: [RegCheck $\rightarrow$ TransactionLog $\rightarrow$ ViolationState]}
}}

\medskip
\noindent$\downarrow$ \textit{Symbolic Verifier $\mathrm{V}(C, C')$}
\medskip

\noindent\fbox{\parbox{0.96\linewidth}{%
\textbf{Fact Extraction:} $\mathcal{F}(C') $ contains \texttt{step: RegCheck} which is \emph{not} in $\mathcal{F}(C)$ \\
\textbf{Soundness:} 7/8 = 0.875 \ding{55} \quad \textbf{Coverage:} 7/8 = 0.875 \quad \textbf{Result: UNSOUND}\\
\textit{The verifier detects that} \texttt{RegCheck} \textit{in the step set is a hallucinated fact---it belongs to the base set in the original certificate. One wrong fact is sufficient to flag the explanation as unsound.}
}}

%% ---- Example 4: Unsound (Entity Hallucination caught by LLM, not by NN) ----
\subsection*{Example 4: LLM Baseline Failure --- Entity Misspelling}
\label{ex:llm_failure}

\noindent\fbox{\parbox{0.96\linewidth}{%
\textbf{Certificate $C$} (NO/bounded\_proof) \\
\texttt{verdict: NO, kind: bounded\_proof} \\
\texttt{checked\_states: \{DerivativePricing, CounterpartyExposure, MarginCallTrigger,}\\
\texttt{\phantom{checked\_states: \{}NettingAgreement, CreditLimitCheck\}}
}}

\medskip
\noindent$\downarrow$ \textit{Best LLM pair (Opus~4.5/Opus~4.5)}
\medskip

\noindent\fbox{\parbox{0.96\linewidth}{%
\textbf{LLM Explanation} \\
The bounded model checking found no violation within the bound. The checked state set includes \texttt{DerivativePricing}, {\color{red}\texttt{CounterPartyExposure}}, \texttt{MarginCallTrigger}, \texttt{NettingAgreement}, and \texttt{CreditLimitCheck}.\\
{\color{red}$\leftarrow$ Misspelling: ``CounterPartyExposure'' (capital P) vs.\ correct ``CounterpartyExposure''}
}}

\medskip
\noindent$\downarrow$ \textit{LLM Reconstruction (Opus~4.5)}
\medskip

\noindent\fbox{\parbox{0.96\linewidth}{%
\textbf{Reconstructed $C'$} \\
\texttt{checked\_states: \{DerivativePricing, {\color{red}CounterPartyExposure}, MarginCallTrigger,}\\
\texttt{\phantom{checked\_states: \{}NettingAgreement, CreditLimitCheck\}}
}}

\medskip
\noindent$\downarrow$ \textit{Symbolic Verifier $\mathrm{V}(C, C')$}
\medskip

\noindent\fbox{\parbox{0.96\linewidth}{%
\textbf{Soundness:} 4/5 = 0.80 \ding{55} \quad \textbf{Coverage:} 4/5 = 0.80 \quad \textbf{Result: UNSOUND}\\
\textit{The LLM generated the state name from its vocabulary rather than copying it, introducing a capitalization error. The symbolic verifier performs exact string matching, correctly flagging} \texttt{CounterPartyExposure} \textit{as a hallucinated entity not present in the original certificate. The pointer-generator mechanism in NN$_1$ prevents this class of error entirely by copying state tokens directly from the source.}
}}

\bigskip
\noindent\textbf{Summary.} These four examples illustrate the key behaviors of the cycle-consistent pipeline: (1)~perfect faithfulness when all facts are correctly copied and attributed; (2)~sound but incomplete explanations that omit facts without hallucinating; (3)~unsound explanations caused by sequence attribution errors (the dominant neural failure mode); and (4)~the LLM baseline's vulnerability to entity-level errors that the pointer-generator mechanism eliminates by design.

\end{document}